\documentclass{article}

\usepackage[utf8]{inputenc}
\usepackage[T1]{fontenc}
\usepackage{hyperref}

\usepackage[accepted]{icml2026}

\usepackage{amsmath}
\usepackage{amssymb}
\usepackage{booktabs}
\usepackage{graphicx}
\usepackage{microtype}
\usepackage{placeins}
\usepackage{float}
\usepackage{array}
\usepackage{url}
\usepackage{xcolor}

\newcommand{\answerTODO}[1][]{\textcolor{red}{\bf [TODO]}}

\graphicspath{{figures_arxiv/}}

\icmltitlerunning{Intertemporal Preference Steering in Qwen3 via CAA}

\begin{document}
\twocolumn[
  \icmltitle{Intertemporal Preference Steering in Qwen3 via Contrastive Activation Addition}

  \begin{icmlauthorlist}
    \icmlauthor{Michal Mráz}{ind}
    \icmlauthor{Justin Shenk}{ind}
  \end{icmlauthorlist}

  \icmlaffiliation{ind}{Independent}

  \icmlcorrespondingauthor{Michal Mráz}{misko.mraz@gmail.com}
  \icmlcorrespondingauthor{Justin Shenk}{shenk.justin@gmail.com}

  \icmlkeywords{Mechanistic Interpretability, Temporal Preference, Activation Steering}

  \vskip 0.3in
]

\printAffiliationsAndNotice{}

\begin{abstract}
We study linear representations of temporal horizon in the large language model
\path{Qwen3-32B} and use them to change the
model's time-related preferences, recommendations, and capabilities. We train contrastive linear probes on
teacher-forced temporal-choice answers to find a short-term versus long-term direction in the model's residual stream and evaluate contrastive activation
addition steering on a held-out binary temporal-choice task, an out-of-distribution monetary intertemporal-choice task, and a
TravelPlanner capability benchmark. The central result is that temporal-horizon
directions can be identified with simple contrastive linear probes and then used
for steering to induce large, bidirectional preference changes. On an
out-of-distribution monetary choice task that varies reward size and delay,
steering strongly shifts the model's indifference threshold between
smaller-sooner and larger-later rewards in both directions. We further show
improvements on a planning-related capability metric under moderate temporal
steering. These results suggest that
model intertemporal preferences are measurable and steerable, which is
relevant for AI systems that give advice involving delayed costs and benefits,
and for safety questions about long-horizon planning.
\end{abstract}

\section{Temporal-Horizon Direction and Probe Diagnostics}
\label{sec:probe-methodology}

\subsection{Direction Construction}

This section shows that intertemporal preference in the residual stream of the large language model \path{Qwen3-32B} is linearly separable and defines the temporal-horizon vector used in the downstream
activation-steering experiments. Linear directions have been used to study
high-level model representations \citep{marks2024geometry}, and contrastive
activation addition uses such directions as steering vectors during inference
\citep{rimsky2024caa}. We therefore focus on a signed contrastive
difference-in-means direction.
The direction is trained on paired short-term and long-term answer continuations.
Because these continuations might also differ in abstraction, urgency, or other
semantic concepts, we treat probe accuracy as evidence that the
training contrast is linearly accessible, not as evidence that the direction
isolates temporal horizon alone. For diagnostic purposes, this direction can be evaluated as a linear probe. We report the long-term versus short-term classification
accuracies and compare them with other standard linear
readouts. We treat the classification accuracies as readout diagnostics but do not necessarily use the method that maximizes classification accuracy for steering.
Causal claims are reserved for the intervention experiments in
Sections~\ref{sec:binary-steering},~\ref{sec:time-utility}, and~\ref{sec:travelplanner}. This separation
is standard in mechanistic interpretability: readout accuracy identifies
accessible information, while activation patching or activation steering tests
whether changing a representation changes behavior
\citep{alain2016linear,tenney2019bert,elhage2021framework,meng2022rome,turner2023activation,rimsky2024caa,li2023iti,zou2023representation}.

All experiments use \path{Qwen3-32B}, with thinking trace disabled \citep{yang2025qwen3,qwen3hf}. Mean-mass (MM) difference
directions and comparison probes are trained at layers
\[
24,\; 28,\; 32,\; 36,\; 40,\; 44,\; 48,
\]
which span the middle-to-late part of the 64-layer model. We use an explicit dataset of
500 temporal-choice questions with overt time markers and an implicit dataset of
300 questions that encode horizon through semantic contrasts without explicit
time words. Both datasets are validated for clear horizon distinction and
balanced presentation order. The explicit set is split at question level into
400 train and 100 test questions. The implicit set is used only for
cross-domain evaluation on the full 300-question set. Each
explicit question yields two teacher-forced examples, so the
explicit training matrix contains 800 examples.

\begin{table*}[t]
\centering
\footnotesize
\setlength{\tabcolsep}{3pt}
\begin{tabular}{>{\raggedright\arraybackslash}p{0.12\textwidth}>{\raggedright\arraybackslash}p{0.30\textwidth}>{\raggedright\arraybackslash}p{0.23\textwidth}>{\raggedright\arraybackslash}p{0.23\textwidth}}
\toprule
Regime & Prompt & Short-term continuation & Long-term continuation \\
\midrule
Explicit &
When developing a roadmap for this initiative, consider: &
What we can deliver in the next 30 days &
What position we want in 10 years \\
Implicit &
When the system fails, the first priority becomes &
containing the spreading damage before it escalates &
redesigning the architecture to prevent recurrence \\
\bottomrule
\end{tabular}
\caption{Representative probe-training examples. Each row yields two
teacher-forced continuations with opposite temporal-horizon labels.}
\label{tab:probe-examples-short}
\end{table*}

The implicit examples are intended to reduce reliance on explicit time words,
but they do not fully isolate horizon: in practice, long-horizon answers often
become more systemic, preventive, and abstract, while short-horizon answers are
often more local, reactive, and operational.

For each training example, the model receives the natural-language question as
context and the short-term and long-term continuations are teacher-forced
as the two distinctive options. The MM difference direction uses
mean-answer pooling, which averages hidden states across all continuation
tokens. Mean-answer pooling is used for the MM difference direction as a design
choice: we expected a span-level answer feature to align better with
interventions applied during generation than a single final-token feature.
Final-token pooling at the last continuation token is also evaluated as
a diagnostic variant. At each selected layer, the MM probe direction is
\[
d_{\mathrm{MM}} = \mu_{\mathrm{long}} - \mu_{\mathrm{imm}},
\]
where each mean is computed from the training activations at a fixed layer and
feature. This contrastive construction follows the broader use of latent
directions as simple behavioral readouts \citep{burns2023latent} and the CAA
practice of adding contrastive residual-stream vectors during inference
\citep{rimsky2024caa}. It is especially relevant here because it preserves a
signed short-term-to-long-term direction that can be added back into the
residual stream at inference.

The same features are also used to fit comparison readouts. Logistic regression
(LR) is fit directly. Whitened logistic regression (WLR) first fits a train-only
covariance whitener, using regularized covariance
\[
\Sigma_{\mathrm{reg}}=\Sigma+10^{-2}\bar{\sigma}^{2}I,
\]
where \(\bar{\sigma}^{2}\) is the average coordinate variance, and then fits LR
in the whitened space. The whitened MM (WMM) probe applies the same covariance
adjustment to the difference-of-means direction. These variants contextualize
linear separability, while the downstream interventions use the unwhitened MM
probe direction as the vector for CAA steering.

\subsection{Probe Diagnostics}

\begin{table*}[t]
\centering
\small
\setlength{\tabcolsep}{4pt}
\begin{tabular}{lcc}
\toprule
Probe & Explicit holdout & Implicit full \\
\midrule
LR  & 99.0\% (L28) & 83.5\% (L24) \\
WLR & 99.0\% (L32) & 80.3\% (L28) \\
MM  & 96.0\% (L24) & 77.0\% (L32) \\
WMM & 99.0\% (L32) & 80.3\% (L28) \\
\bottomrule
\end{tabular}
\caption{Explicit-trained probe results for the main mean-answer pooling
feature. Entries report the best accuracy and the layer attaining it. The MM
row is the probe family from which the CAA steering vector is constructed.}
\label{tab:explicit-probe-mean}
\end{table*}

\begin{figure*}[t]
\centering
\includegraphics[width=0.92\textwidth]{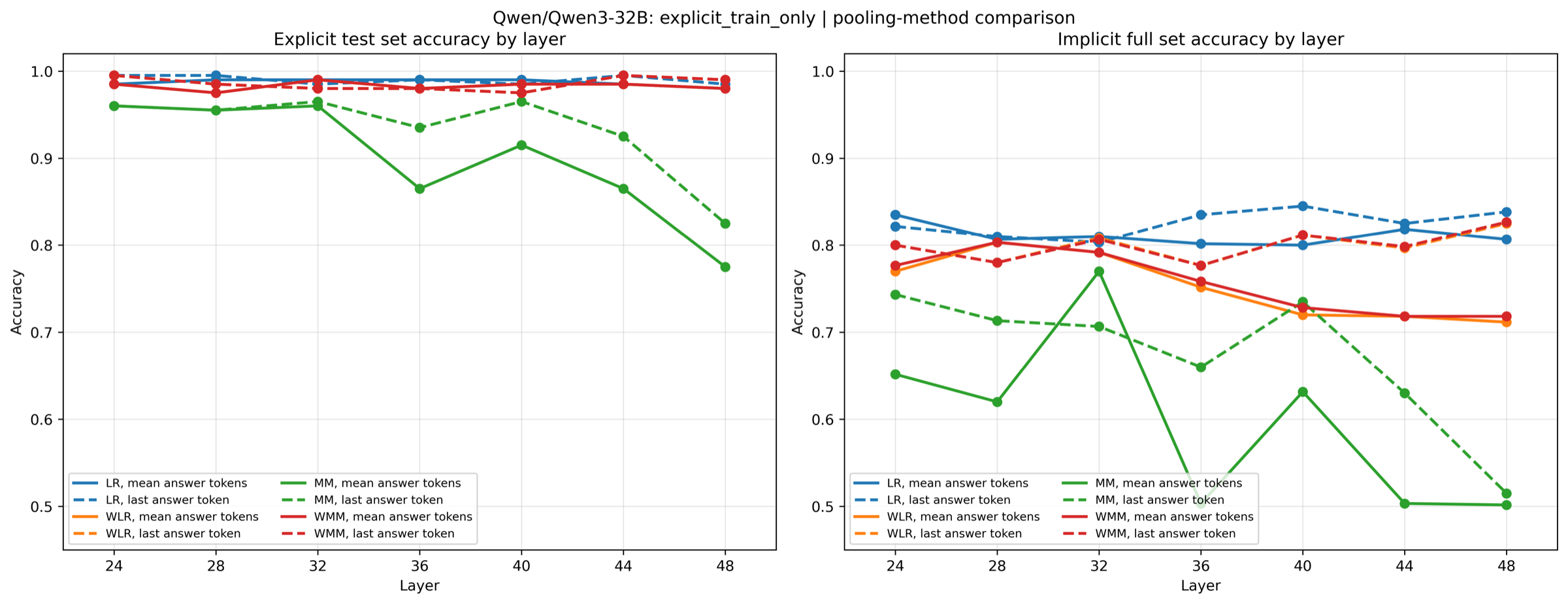}
\caption{Layerwise explicit-trained probe accuracies from the probe-variation
experiment. The left panel reports explicit test set accuracy; the
right panel reports full implicit set accuracy.}
\label{fig:explicit-probe-comparison}
\end{figure*}

All probe families nearly saturate explicit holdout accuracy, while cross-domain
implicit performance is lower but still substantial. The MM row shows that the
specific contrastive direction used for CAA steering remains a strong temporal
horizon readout even though LR and whitened variants attain slightly higher
classification accuracy. Notably, the two whitened probe families have very
similar classification accuracies across the reported splits. The
steering experiments therefore use CAA steering with the explicit-trained
mean-answer MM difference direction.

\section{Binary Temporal Steering}
\label{sec:binary-steering}

This section shows that CAA steering with the explicit-trained MM difference
direction causally affects temporal-choice behavior on held-out binary
choices from the same broad short-term/long-term answer family. The evaluation uses similarly structured
but unseen binary-choice questions. No evaluation question is used to fit the
probe direction, and the only intervention is a signed residual-stream addition
during inference. If positive and negative additions produce systematic changes
in both parsed choices and forced-choice log probabilities, the direction is not
only a classifier direction but a causal control direction for this task family.
The intervention uses the explicit-trained mean-answer MM difference direction from
Section~\ref{sec:probe-methodology}.

\subsection{Evaluation Prompt}
Unlike probe training, steering evaluation presents the full binary-choice
context, so the model must choose between two complete answer candidates rather
than score an answer in isolation:
\begin{flushleft}
\small\ttfamily
\begin{tabular}{@{}l@{}}
<question>\\
Options:\\
\ \ <short-term or long-term option>\\
\ \ <the other horizon option>\\
Answer:
\end{tabular}
\end{flushleft}
The short-term and long-term options are presented in randomized order, and the
evaluated datasets are the held-out explicit test set with 100 prompts and the
full implicit set with 300 prompts.

\subsection{Intervention}
For a selected layer \(l\), signed strength \(\alpha\), and normalized MM difference vector
\(\hat d_l\), the hook adds \(\alpha \hat d_l\) to the residual stream at the
last prompt token and the same update is also applied at each decode step.
Positive strengths steer toward the long-term class and negative strengths steer
toward the immediate class. Table~\ref{tab:binary-steering-config} summarizes
the intervention grid.

\begin{table*}[t]
\centering
\small
\setlength{\tabcolsep}{5pt}
\begin{tabular}{p{0.26\textwidth}p{0.58\textwidth}}
\toprule
Quantity & Values \\
\midrule
Steered layers \(l\) & \(24,\;28,\;32,\;36,\;40,\;44,\;48\) \\
Signed strengths \(\alpha\) & \(0,\;\pm2,\;\pm4,\;\pm8,\;\pm16,\;\pm32,\;\pm64,\;\pm128\) \\
Patch positions & Last prompt token and every generated-token step \\
\bottomrule
\end{tabular}
\caption{Binary steering configuration. Each intervention adds
\(\alpha\hat d_l\) to the residual stream at layer \(l\).}
\label{tab:binary-steering-config}
\end{table*}
We do not claim a uniquely optimal layer. We treat layers 24--48 as an
exploratory probe-layer band and report full sweeps where feasible. The main
result is that CAA effects are broad across middle-to-late layers, not that
any specific layer is intrinsically privileged. Single-layer plots are
representative slices chosen for readability or compute constraints.

Because the steering vectors are normalized, the added residual perturbation has
L2 norm \(|\alpha|\). The mean hook-site residual L2 norms in baseline (unsteered) 
generations are between 202 at layer 24 and 355 at layer 48. The
strongest intervention \(|\alpha|=128\)
is therefore 36\%--63\% of the unsteered norm, which is a deliberately large perturbation, 
while \(|\alpha|\le16\) is less than 8\% of the corresponding residual norm.
For details on intervention and baseline residual stream magnitudes, see Appendix~\ref{app:residual-norms}.

\subsection{Choice and Likelihood Metrics}
Generated continuations are parsed into long-term or immediate choices, and the
main behavioral metric is the proportion of parsed continuations choosing the
long-term option. If the parser cannot identify the choice, the evaluator falls
back to teacher-forced candidate logprob comparison; the plotted choice rates
are computed from parsed, no-fallback choices. The likelihood
metric computes
\[
\begin{aligned}
\Delta_{\mathrm{avg}}
&= \ell(c_{\mathrm{long}}\mid p)-\ell(c_{\mathrm{imm}}\mid p),\\
\ell(c\mid p)
&= \frac{1}{|c|}\sum_t \log P(c_t\mid p,c_{<t}),
\end{aligned}
\]
and reports its change relative to the unsteered baseline. This average-token
logprob margin controls for continuation length; the full layerwise logprob
surface is shown in Appendix~\ref{app:binary-steering-logprob},
Figure~\ref{fig:binary-steering-logprob}.

\begin{figure*}[t]
\centering
\includegraphics[width=0.92\textwidth]{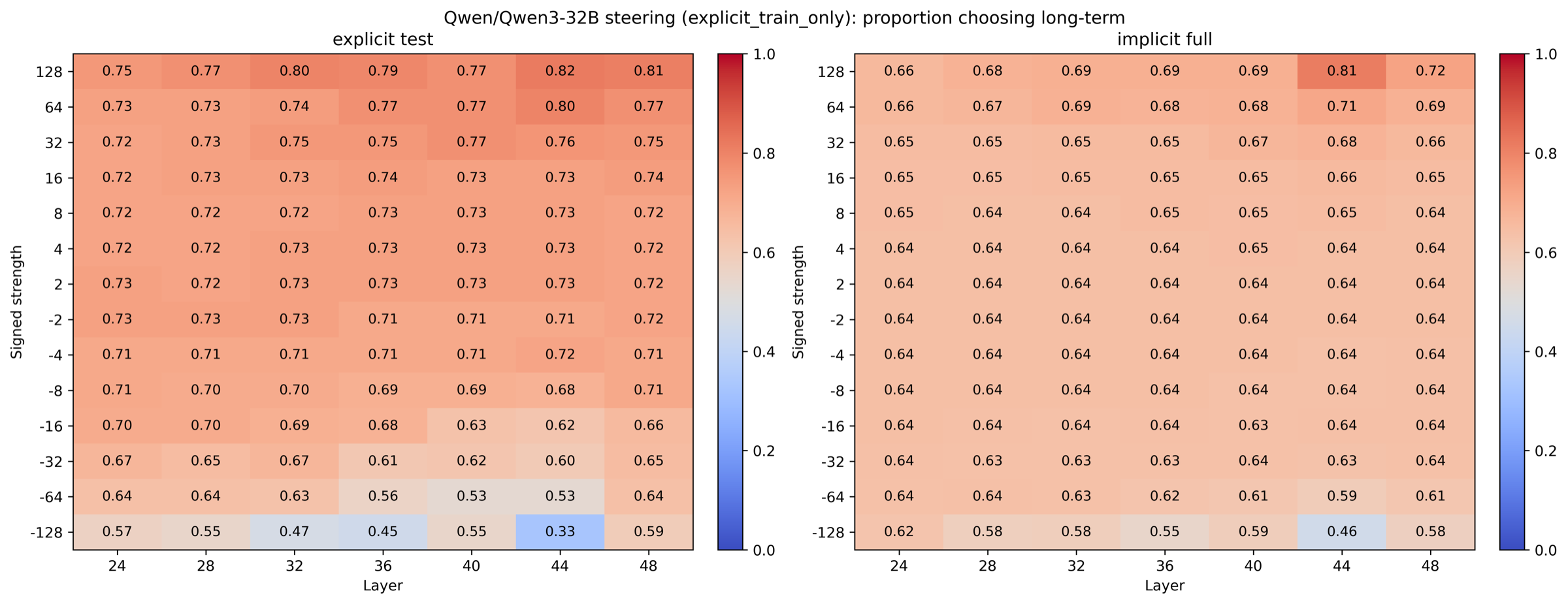}
\caption{Proportion of prompts choosing the long-term option over layer and
signed steering strength.}
\label{fig:binary-steering-choice}
\end{figure*}

We see the causal steering effect across all explored layers, with layer 44
giving the largest parsed-choice contrast on both evaluation sets. This broad
middle-to-late layer profile is consistent with prior activation-steering and
intervention work, which often finds the strongest causal directions away from
the earliest and latest layers of the stack
\citep{turner2023activation,rimsky2024caa,durmus2024steering}. Note that these numbers only describe the
measured intervention profile; they do not by themselves identify a complete
circuit for intertemporal preference.

As a matched-norm control, we also steer with a random unit vector orthogonal to
the MM difference direction at each layer. This control produces substantially smaller
choice shifts than CAA steering, and those shifts are not consistently aligned
with the sign of the intervention (Appendix~\ref{app:binary-random-control},
Figure~\ref{fig:binary-random-orthogonal-control}).
The main binary results use greedy decoding with temperature \(0\), so repeated
decoding is not a source of sampling variance for those point estimates. As a
robustness check, however, Appendix~\ref{app:binary-stochastic} repeats a stochastic
layer sweep with temperature \(0.8\), top-\(p=1\), and five sampled generations
per prompt. The sampled means preserve the greedy ordering and magnitudes, with
2-sigma standard-error bars small relative to the steering effect.

\section{Time-Utility Steering}
\label{sec:time-utility}

This section shows generalization beyond the binary temporal-choice distribution
toward settings where models might advise users about real-world tradeoffs.
The model is asked monetary time-utility questions involving dollar amounts and
delays, a quantitatively measurable task that was not used to train the temporal
horizon direction. The goal is to show that a direction trained from semantic
short-term versus long-term answers can change preferences on a structurally
distinct advice task, not merely on held-out examples with the same format. This
setting is related to classical time preference and discounting
measurements \citep{samuelson1937utility,frederick2002time}.

\subsection{Task and Steering Setup}
\label{sec:time-utility-setup}

The prompt is written from the user's point of view:
\begin{quote}
\small\ttfamily
\begin{tabular}{@{}l@{}}
I have an offer to take <x>\$ now\\
or <y>\$ in <t>.\\
What should I choose?\\
Answer in just a few words.
\end{tabular}
\end{quote}
The monetary grid is defined by a base immediate amount \(x_b\), base delayed
amounts \(y_b\), and amount scale factors \(c\). Each prompt uses
\(x=c x_b\) and \(y=c y_b\), so the delayed/immediate multiplier
\(m_b=y_b/x_b\) is held fixed while the absolute dollar amounts vary. This
separates numeric proportions from absolute-dollar lexical effects. The
experiment also uses day-count delay labels where possible rather than mixed natural labels.
For example, ``in 365 days'' and ``in a year'' denote approximately the same
delay, but the latter introduces additional calendar semantics.

\begin{table*}[t]
\centering
\small
\setlength{\tabcolsep}{5pt}
\begin{tabular}{p{0.25\textwidth}p{0.60\textwidth}}
\toprule
Quantity & Values \\
\midrule
Base immediate amount \(x_b\) & \(100\) \\
Base delayed amounts \(y_b\) & \begin{tabular}[t]{@{}l@{}}
\(100,\;101,\;110,\;150,\;200,\;500,\;1000,\;10000\)
\end{tabular} \\
Multipliers \(m_b=y_b/x_b\) & \begin{tabular}[t]{@{}l@{}}
\(1.0,\;1.01,\;1.1,\;1.5,\;2,\;5,\;10,\;100\)
\end{tabular} \\
Amount scale factors \(c\) & \(0.3,\;0.5,\;0.7,\;1,\;1.4,\;2,\;6,\;8\) \\
Delays \(t\) & \begin{tabular}[t]{@{}l@{}}
\texttt{1 hour}, \texttt{8 hours}, \texttt{1 day}, \texttt{2 days}, \texttt{7 days}, \texttt{30 days},\\
\texttt{90 days}, \texttt{365 days}, \texttt{730 days}, \texttt{1095 days}, \texttt{1825 days}, \texttt{3650 days}
\end{tabular} \\
Signed strengths \(\alpha\) & \(0,\;\pm8,\;\pm16,\;\pm32,\;\pm64,\;\pm128\) \\
\bottomrule
\end{tabular}
\caption{Time-utility steering grid. For each \(y_b\), scale factor \(c\), and
delay \(t\), the prompt compares \(x=cx_b\) now against \(y=cy_b\) after
delay \(t\).}
\label{tab:time-utility-grid}
\end{table*}

Each condition therefore has \(8\) amount multipliers, \(12\) delays, and \(8\)
scale factors, for \(768\) generations.
Parsed responses are categorized as choosing the immediate amount, choosing the
delayed amount or ``it depends''.

\subsection{Area-Equivalent Indifference Multiplier}
\label{sec:time-utility-aeim}

The area-equivalent indifference multiplier estimates the delayed/immediate
reward ratio \(y/x\) at which the model is indifferent between taking \(x\) now
and \(y\) after delay \(t\). For a fixed condition and delay, all scale-factor
rows with the same base
multiplier \(m=y/x\) are combined. Let \(q_{\mathrm{now}}(m,t)\) be the parsed
present-choice fraction at multiplier \(m\) and delay \(t\). ``It depends''
responses are split equally between present and delayed choices, although this convention
has negligible effect because ``it depends'' occurs in only 0.15\% of responses. 
Write \(z_i=\log m_i\) for the sorted sampled multipliers in a delay column and define
midpoint bin edges on the log axis by \(e_0=z_{\min}\), \(e_n=z_{\max}\), and
\(e_i=\tfrac12(z_i+z_{i+1})\) for \(1\le i<n\). Each sampled multiplier is then
assigned bin width \(w_i=e_i-e_{i-1}\). Treating
\(q_{\mathrm{now}}(z,t)\) as piecewise constant on these bins, the ideal
full-coverage log-axis average is
\[
A_{\mathrm{now}}(t)=
\frac{\sum_i w_i q_{\mathrm{now}}(m_i,t)}{\sum_i w_i},
\]
since \(\sum_i w_i=z_{\max}-z_{\min}\), the log indifference multiplier
\(z^*(t)\) is
\[
\begin{aligned}
z^*(t)
&= z_{\min}+A_{\mathrm{now}}(t)(z_{\max}-z_{\min})\\
&= z_{\min}+\sum_i w_i q_{\mathrm{now}}(m_i,t).
\end{aligned}
\]
This is a piecewise-constant log-axis average, chosen because the tested
multipliers are sparse and unevenly spaced and the empirical choice columns can
be locally non-monotone. The corresponding
indifference multiplier on the original scale is \(m^*(t)\):
\[
m^*(t)=\exp(z^*(t)).
\]
Large \(m^*(t)\) means that the model requires a larger delayed reward before it
switches away from the immediate option, while small \(m^*(t)\) means that a
smaller delayed reward is sufficient to make the model wait.

Appendix~\ref{app:time-utility-details} shows the full present-choice heatmaps
and precise AEIM values. The heatmaps show the same directional pattern summarized by AEIM: negative steering
expands the present-choice region, while positive steering contracts it, without
relying on the AEIM aggregation.

\subsection{Layer Selection and Results}
\label{sec:time-utility-results}

\begin{figure*}[t]
\centering
\includegraphics[width=0.78\textwidth]{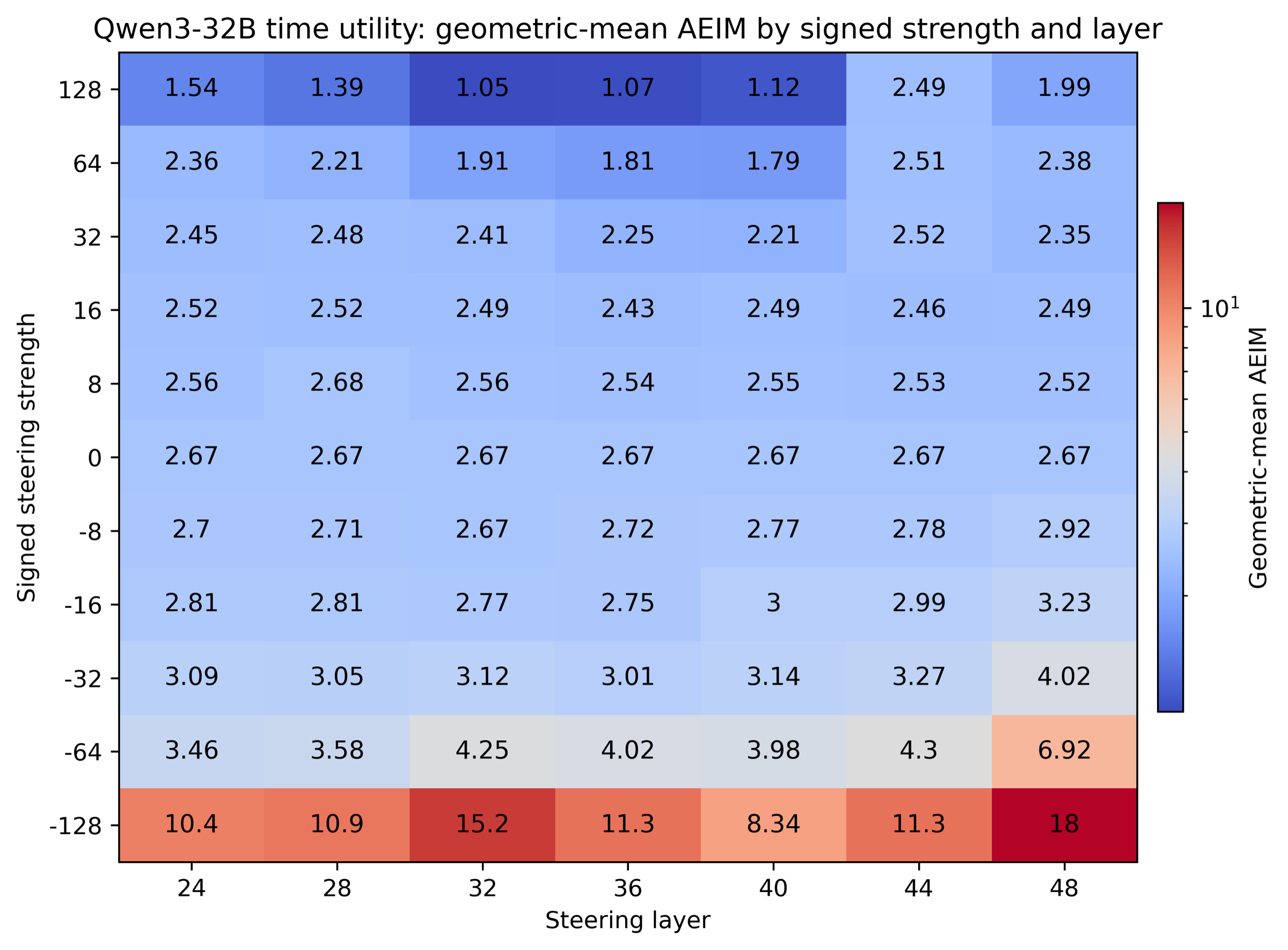}
\caption{Time-utility layer sweep. Each cell reports the geometric mean AEIM.}
\label{fig:time-utility-layer-sweep}
\end{figure*}

We sweep the same seven probe layers as in the binary-steering experiment.
Figure~\ref{fig:time-utility-layer-sweep}
summarizes the geometric mean AEIM over the 12 delays for each layer and signed
strength. The effect is signed and broad across layers. The layer-40 curves below are therefore an illustrative slice
of the sweep, not the unique selected optimum. The other layerwise AEIM curves
are shown in Appendix~\ref{app:time-utility-details}.

\begin{figure*}[t]
\centering
\includegraphics[width=0.92\textwidth]{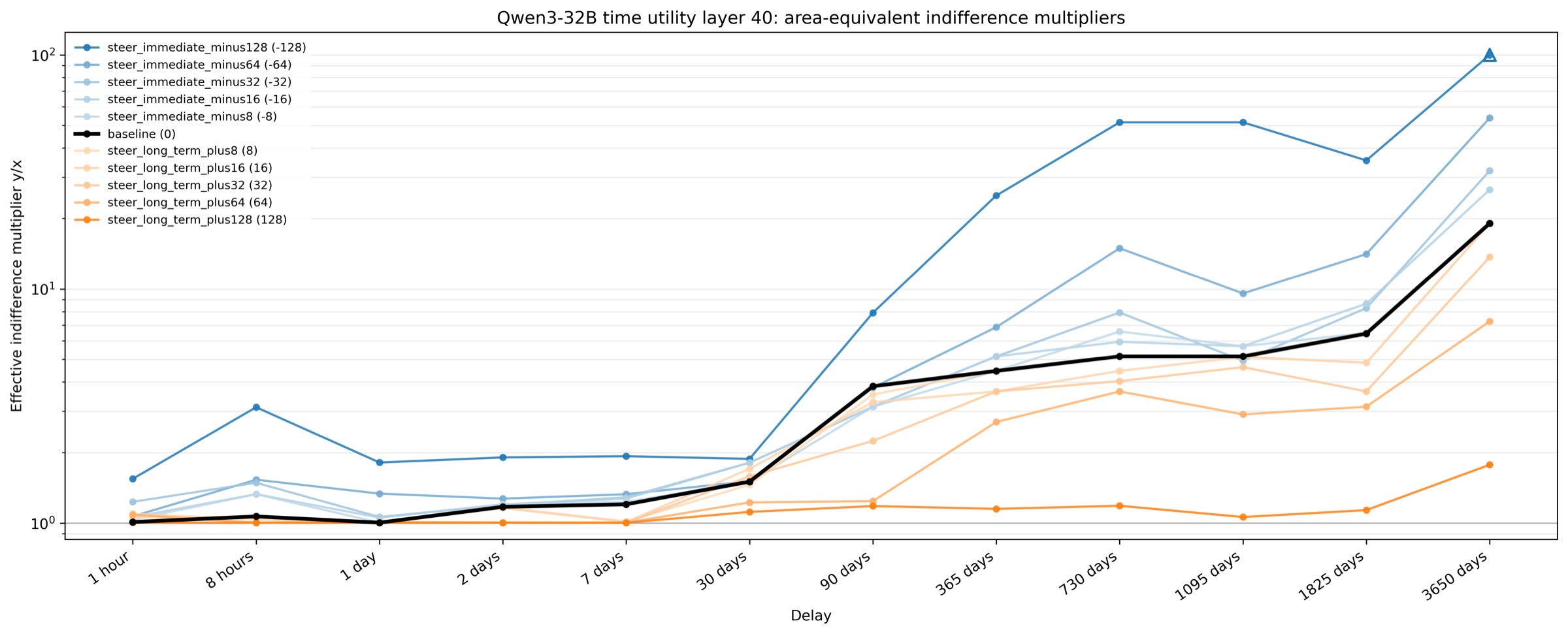}
\caption{Area-equivalent indifference multipliers from the layer-40
time-utility experiment. Note the -128 steering at 3650 days is capped at 100 because all model responses chose the immediate option.}
\label{fig:time-utility-aeim}
\end{figure*}

\begin{table*}[t]
\centering
\small
\setlength{\tabcolsep}{4pt}
\begin{tabular}{lcccc}
\toprule
Residual addition & Geom. mean AEIM & Parsed later rate & It-depends rate & Unparsed rate \\
\midrule
\(-128\hat d_{40}\) & 8.34 & 36.3\% & 0.0\% & 0.26\% \\
0 & 2.67 & 54.7\% & 0.0\% & 0.00\% \\
\(+128\hat d_{40}\) & 1.12 & 80.4\% & 1.2\% & 0.52\% \\
\bottomrule
\end{tabular}
\caption{Layer-40 aggregate time-utility results over all delay and amount cells. The
geometric mean AEIM is computed across the 12 delay columns. Parsed later and
it-depends rates are computed over parsed generations in the condition.}
\label{tab:time-utility-aggregate}
\end{table*}

The AEIM curves show that the result is delay-dependent rather than uniform:
as expected, negative steering moves the indifference multiplier upward, positive
steering moves it downward, and most displacement occurs at long delays. At \(\pm128\),
the median ratio of negative-steer AEIM to positive-steer AEIM across delays is
4.90 and at 3650 days it is over 56. In other words, at the
10-year horizon, negative steering makes the model require more than
56 times the delayed reward needed under positive steering. The result is a
strong generalization test: a direction trained only from binary temporal-horizon
answers changes the model's measured intertemporal preferences on a new
monetary-choice task.

For robustness, we performed a matched random-orthogonal control, which produces smaller, inconsistently signed
movement (Appendix~\ref{app:time-utility-details}). Unlike the MM difference vector, the
random vector does not produce a consistent signed short-term versus long-term
pattern across layers.

The main sweep uses greedy decoding with temperature \(0\). Analogously to
Section~\ref{sec:binary-steering}, Appendix~\ref{app:time-utility-stochastic}
reports robustness checks with stochastic sampling, which preserve the same
qualitative pattern. Appendix~\ref{app:residual-norms} reports the steering
vector magnitudes relative to unsteered hook-site residual-stream magnitudes,
which are always less than 70\% even at the strongest intervention.

\section{TravelPlanner Capability}
\label{sec:travelplanner}

This section examines how temporal-horizon steering affects a
planning-relevant capability metric, rather than only changing expressed
preferences. We evaluate the same layer-40 temporal direction on TravelPlanner,
a real-world travel planning benchmark that requires models to produce multi-day
itineraries satisfying transportation, accommodation, restaurant, attraction,
budget, and constraint requirements \citep{xie2024travelplanner}. A planning
benchmark is a natural test for long-horizon planning because it requires
maintaining a temporally extended plan while satisfying many local constraints.
We use the validation split: 180 queries, comprising 20 queries for each
combination of three difficulty levels and three trip lengths. Each query asks
for a complete plan containing transportation, accommodation, meals, and
attractions for each day.
We use layer 40 as a compute-constrained representative layer from the
middle-to-late probe band rather than as a task-specific optimum.

\subsection{Metric Choice}
The benchmark has multiple metrics; we use Commonsense Constraint Micro Pass
Rate as the primary metric for this experiment. It is the total number of passed commonsense checks divided by the total
number of commonsense checks across all evaluated queries. Commonsense micro
pass rate is the most informative metric for our purposes. In the baseline
condition (no steering), delivery rate is already near ceiling (96.7\%), while
commonsense macro pass rate, hard-constraint rates, and final pass rate are near
floor. Commonsense micro pass rate therefore captures partial improvements in
itinerary plausibility even when strict whole-query metrics remain too coarse to
distinguish conditions.

\begin{figure*}[t]
\centering
\includegraphics[width=0.82\textwidth]{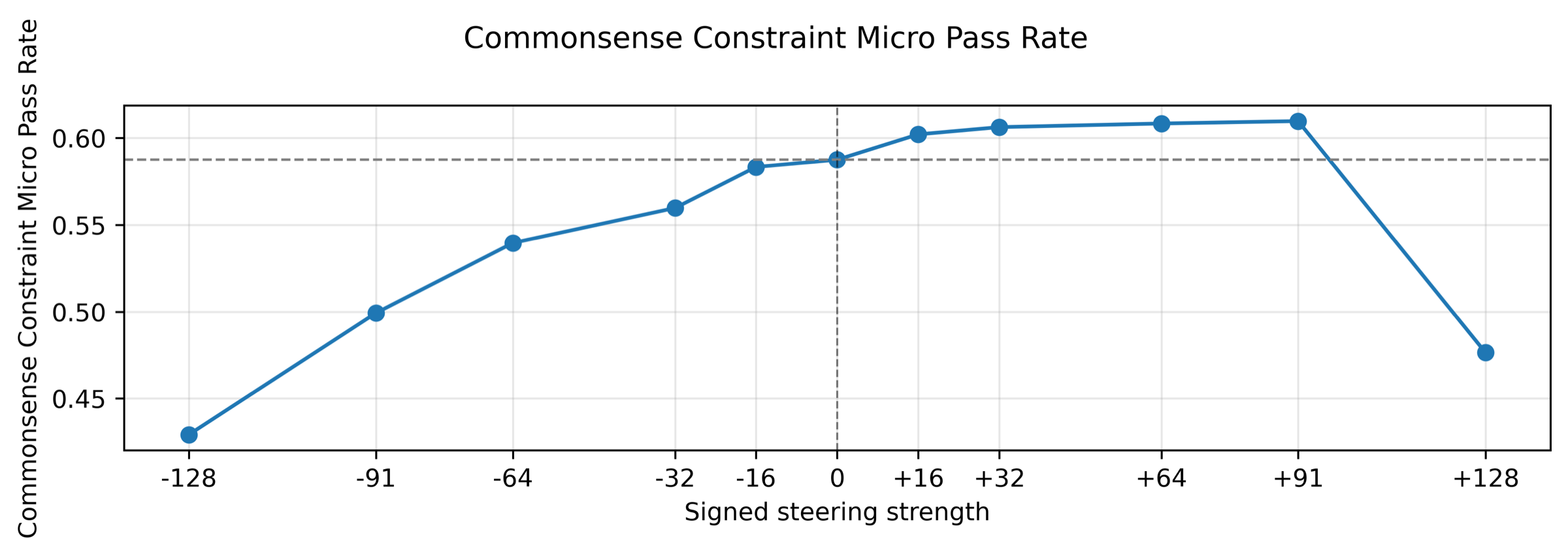}
\caption{TravelPlanner Commonsense Constraint Micro Pass Rate as a function of
signed temporal steering strength. Positive values steer toward the long-term
direction; negative values steer toward the short-term direction.}
\label{fig:travelplanner-commonsense-micro}
\end{figure*}

Figure~\ref{fig:travelplanner-commonsense-micro} shows a non-monotonic but
directionally meaningful effect. Moderate long-term steering improves
commonsense micro pass rate, while negative steering degrades it. Very large
positive steering also degrades performance. A likely explanation is that
\(+128\) pushes the model far enough off-distribution that generation coherence
begins to deteriorate. The result is not evidence that temporal steering
generally solves planning, and we do not know from this aggregate alone which
commonsense checks flip under steering, but it suggests that the same direction that shifts intertemporal
preferences can also affect a planning-relevant capability metric, with useful
effects concentrated at moderate positive strengths.

\section{Discussion}
\label{sec:Discussion}

\subsection{Summary}

These experiments suggest that temporal horizon can be represented by simple
linear directions in Qwen3 and that CAA steering with a direction trained from
contrastive short-term versus long-term answers can produce large behavioral
effects. The direction should not be interpreted as an isolated
temporal-horizon feature, since the training contrast may also contain
correlated semantic features. Nevertheless, the time-utility experiment
suggests that intertemporal preference is at least a relevant component of the
direction. The probe results establish that
short-term and long-term answers are linearly separable, and the binary steering
experiment shows that steering along the direction causally affects held-out temporal-choice
questions. The time-utility experiment is the strongest generalization result.
A direction trained only on binary temporal-horizon answers shifts monetary
intertemporal choice on a structurally different task, moving AEIM by large
factors in both directions. The TravelPlanner experiment further suggests that
temporal-horizon steering can affect a planning-relevant capability metric,
although the effect is moderate and non-monotonic.

The main takeaway is that MM difference directions built from contrastive
temporal-horizon answers can generalize beyond their training format when used
for CAA steering. This matters because many
deployed model outputs implicitly encode
intertemporal preferences, such as advice about planning, finance, health,
education, and policy that trades near-term actions against delayed outcomes.
The results suggest that such intertemporal preferences can be measured and
steered, rather than treated as an opaque side effect of pretraining or
instruction tuning. This is also safety-relevant because temporal preferences
may be one ingredient in long-horizon planning and reward-seeking, and a model
that internally optimizes over a different horizon than the one it presents to
users may be difficult to evaluate from outputs alone.
At the same time, the capability to steer temporal preferences could be abused
by a malicious actor to push a model toward decisions that are misaligned with
user values, so the results underscore the importance of secure access
controls and monitoring for similar behavior.

\subsection{Limitations}
\label{sec:limitations}

The results rely on several assumptions that may fail in practice:

\textbf{Construct validity.} The MM direction is trained on
    short-term versus long-term answer continuations, but those continuations
    can also vary in abstraction, urgency, strategic scope, and
    proactive-versus-reactive framing. If these correlated features dominate the
    direction, the intervention should be interpreted as steering a broader
    short-term/long-term answer register rather than an isolated temporal-horizon
    variable. The monetary time-utility transfer suggests intertemporal
    preference is involved, but it does not fully identify the direction's
    semantic basis.

\textbf{Intervention scale and specificity.} Large activation
    additions may change more than temporal preference. The residual-norm
    diagnostics calibrate intervention size at the hook sites, but they do not
    prove that unrelated capabilities are preserved, especially at the strongest
    steering values. In deployment, an intervention that shifts time preference
    while also degrading factuality, instruction following, or coherence would
    be less useful and potentially unsafe.

\textbf{Scope of empirical claims.} All experiments use one model
    family and size, a small set of prompt templates, and thinking-disabled
    decoding. The results may not transfer to other architectures, smaller or
    larger models, reasoning-enabled settings, or tasks where temporal
    tradeoffs are expressed less explicitly. Performance may also depend on
    prompt wording or answer-format instructions.

\textbf{Layer, strength, and prompt sensitivity.} Layer and strength
    choices are exploratory. Full sweeps are reported where feasible, but some
    downstream experiments use representative layers for compute reasons.
    Performance may depend on layer choice, steering norm, and decoding
    settings.

\textbf{Prompting baselines.} The paper studies activation steering,
    not the full space of ways to alter model behavior. Similar behavioral
    changes might be possible by prompting the model to think more short-term or
    long-term, and such baselines are not reported here.

\textbf{Parsing imprecision.} Some generated answers are unparsed,
    especially in the free-generation steering experiments. The unparsed
    fraction is small enough that it is unlikely to change the qualitative
    results, but it can still introduce imprecision into parsed choice rates and
    derived quantities.

\textbf{AEIM aggregation.} AEIM compresses a sparse, unevenly sampled
    amount-delay grid into one threshold-like curve per condition. Different
    multiplier grids, delay grids, or aggregation conventions could change the
    apparent curve shape. The raw present-choice heatmaps in Appendix~\ref{app:time-utility-details}
    show the same steering pattern without this aggregation, but they have a
    less clean scalar interpretation.

\textbf{Probe-family comparisons.} Steering is evaluated
    primarily with the MM difference direction. Matched-norm comparisons against
    LR, WLR, and WMM steering directions would clarify whether MM is uniquely
    useful for intervention.

\textbf{Benchmark coverage.} The TravelPlanner evidence is weaker than the
    time-utility result because the aggregate commonsense metric is indirect and
    does not identify which item-level checks change under steering.

\subsection{Compute Resources}
\label{sec:compute-resources}

All experiments used Qwen3-32B inference.
Activation collection and steering runs were executed on NVIDIA GPU hardware
with 100 GB of available VRAM on RTX 5090 or similar-capability hardware.
The layer-sweep steering experiments took approximately 48 hours for
binary-choice steering and 24 hours for time-utility steering. The respective
stochastic steering sweeps took about twice as long: 4 days and 2 days.
TravelPlanner evaluation with single-layer steering took approximately 48 hours.
The remaining diagnostics and plotting took seconds to minutes on a MacBook Air M4.

\section*{Acknowledgements}

We thank Avigya Paudel for his help with the dataset curation and Joseph
Rudoler for his thoughtful insights during our discussions. Finally, we thank
SPAR for providing the compute resources for the research.

\clearpage
\bibliographystyle{icml2026}
\bibliography{qwen3_32b_temporal_steering_refs}

\clearpage
\appendix
\section*{Appendix}

\section{Binary Steering Layerwise Summary}
\label{app:binary-steering-diagnostics}

Table~\ref{tab:binary-steering-layer-gaps} gives the
layerwise contrast table discussed in Section~\ref{sec:binary-steering}. 
Layer 48 is marginally largest by logprob
gap on both datasets, while layer 44 is largest by choice gap.

\begin{table*}[t]
\centering
\small
\setlength{\tabcolsep}{4pt}
\begin{tabular}{ccccc}
\toprule
Layer & Explicit choice gap & Explicit logprob gap
& Implicit choice gap & Implicit logprob gap \\
\midrule
24 & 23.9 pp & 0.158 & 4.6 pp  & 0.112 \\
28 & 25.3 pp & 0.125 & 10.2 pp & 0.121 \\
32 & 37.6 pp & 0.175 & 11.0 pp & 0.101 \\
36 & 37.5 pp & 0.183 & 14.6 pp & 0.084 \\
40 & 28.0 pp & 0.399 & 10.0 pp & 0.146 \\
44 & 59.4 pp & 0.813 & 36.5 pp & 0.263 \\
48 & 25.7 pp & 0.824 & 14.6 pp & 0.268 \\
\bottomrule
\end{tabular}
\caption{Layerwise steering contrast. The choice gap is the best positive
long-term no-fallback rate minus the best negative long-term no-fallback rate at
that layer. The logprob gap is the best positive minus best negative
long-minus-immediate average-logprob delta at that layer.}
\label{tab:binary-steering-layer-gaps}
\end{table*}

\section{Binary Steering Logprob Diagnostics}
\label{app:binary-steering-logprob}

\begin{figure*}[t]
\centering
\includegraphics[width=0.92\textwidth]{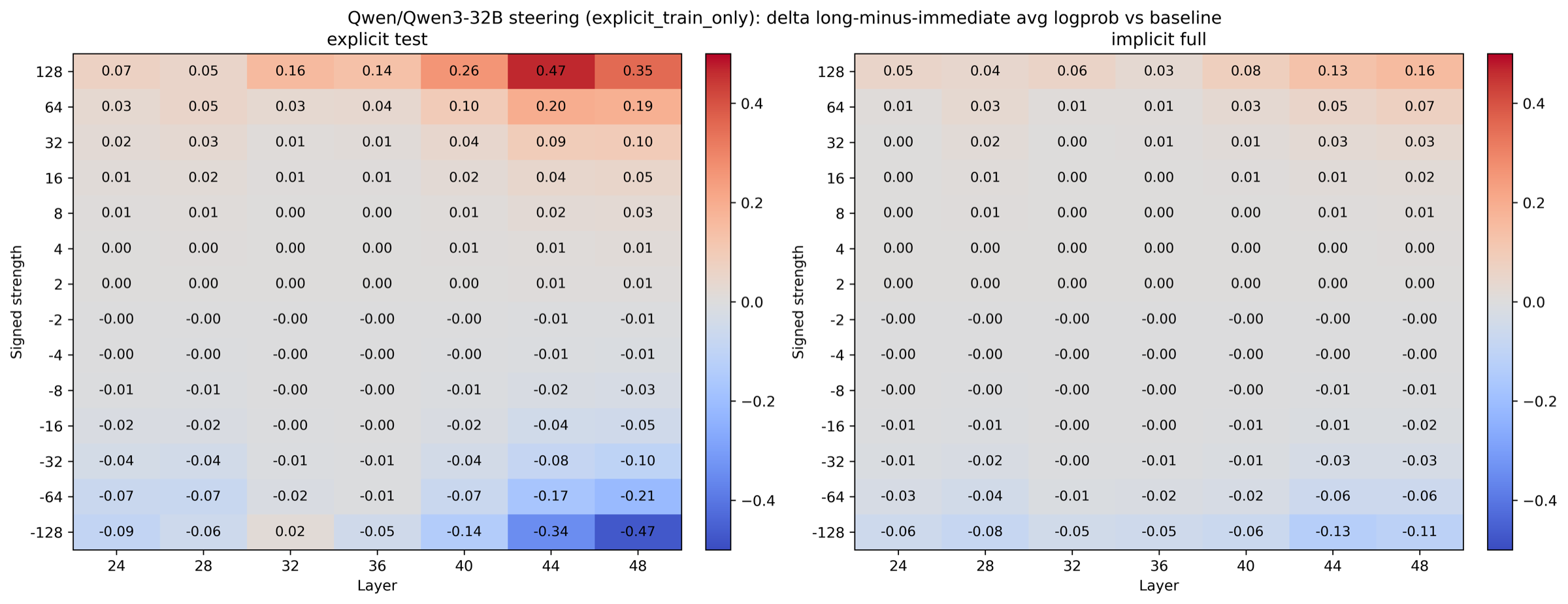}
\caption{Change in long-minus-immediate average logprob margin relative to the
unsteered baseline.}
\label{fig:binary-steering-logprob}
\end{figure*}

\section{Binary Random-Orthogonal Control}
\label{app:binary-random-control}

\begin{figure*}[t]
\centering
\includegraphics[width=0.92\textwidth]{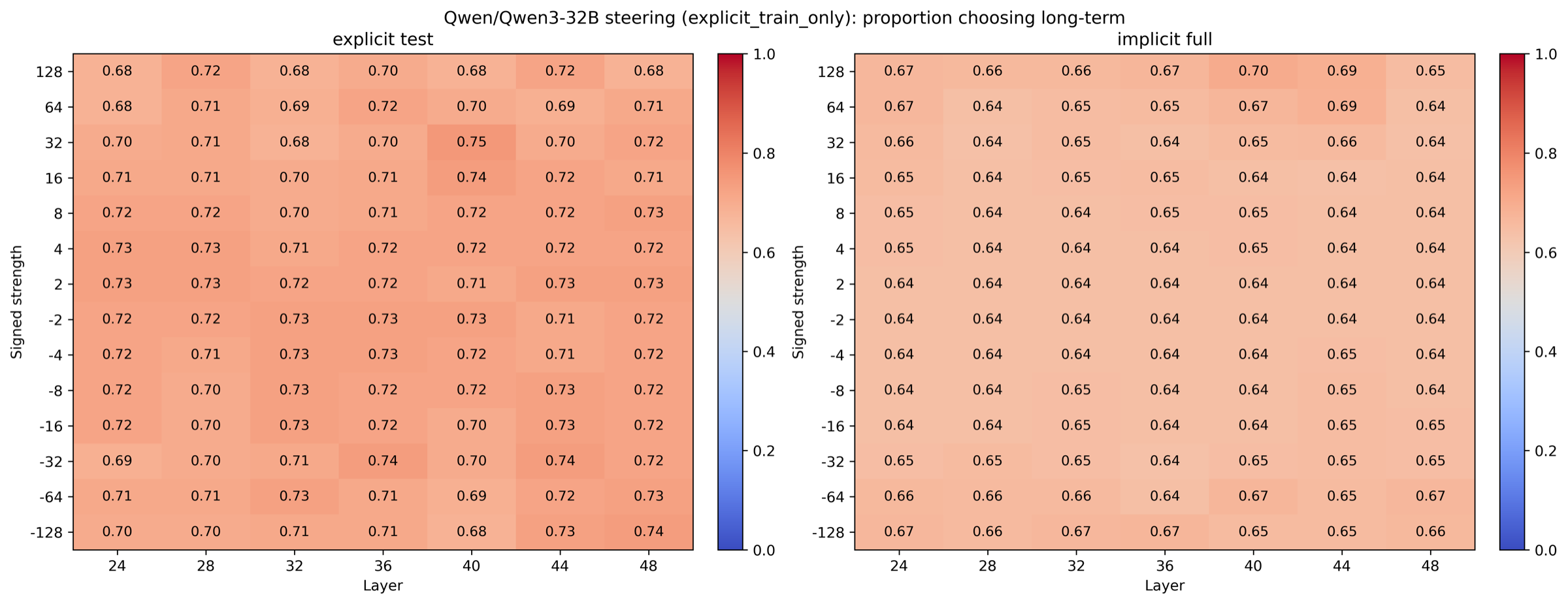}
\includegraphics[width=0.92\textwidth]{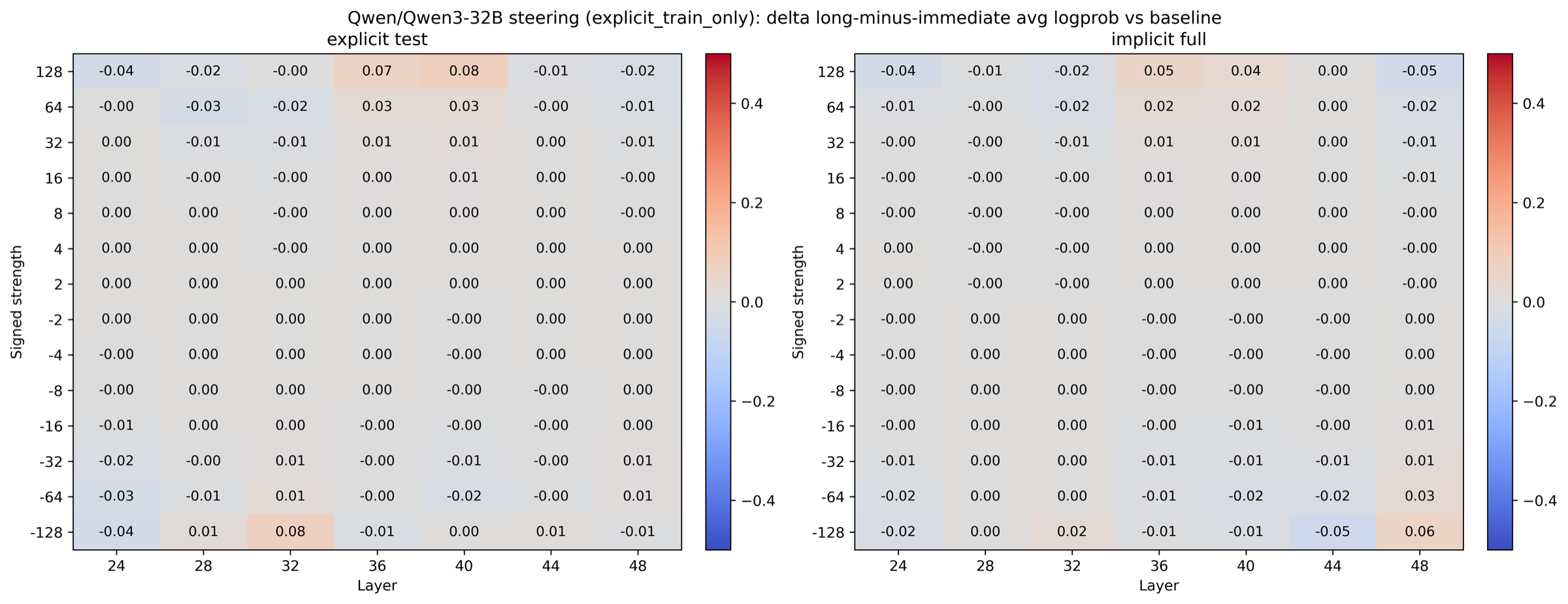}
\caption{Matched random-orthogonal binary steering control. The control vector
at each layer is orthogonalized against the corresponding normalized MM
difference vector.
The largest absolute no-fallback choice movement is 5.6 percentage points on
explicit test and 5.5 points on implicit full, compared with 59.4 and 36.5
percentage-point CAA contrasts for layer-44 steering on the same datasets.}
\label{fig:binary-random-orthogonal-control}
\end{figure*}

\section{Stochastic Binary Steering Robustness}
\label{app:binary-stochastic}

\begin{figure*}[t]
\centering
\includegraphics[width=0.92\textwidth]{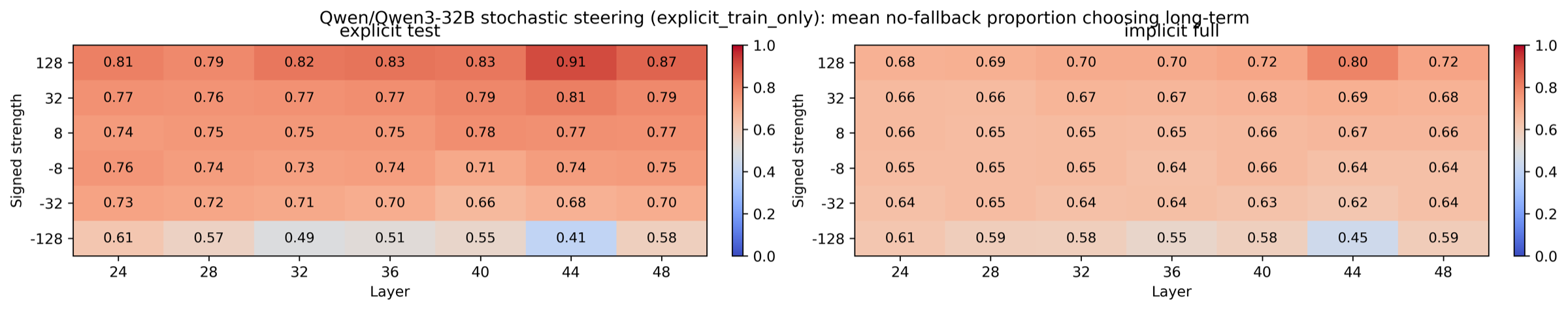}
\includegraphics[width=0.92\textwidth]{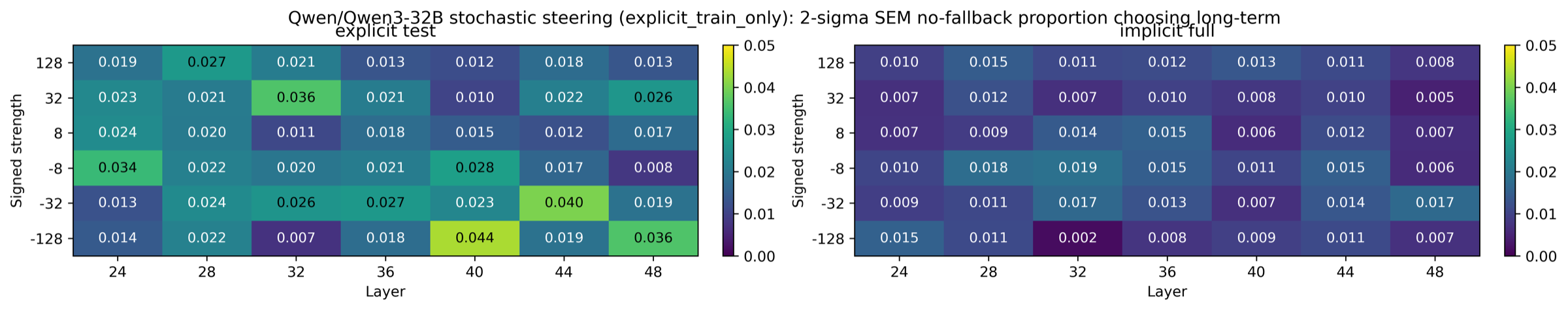}
\caption{Stochastic binary steering no-fallback choice results. Top: mean
proportion choosing the long-term option. Bottom: 2-sigma standard-error bar
across five sampled decodings.}
\label{fig:binary-stochastic-choice}
\end{figure*}

\begin{figure*}[t]
\centering
\includegraphics[width=0.92\textwidth]{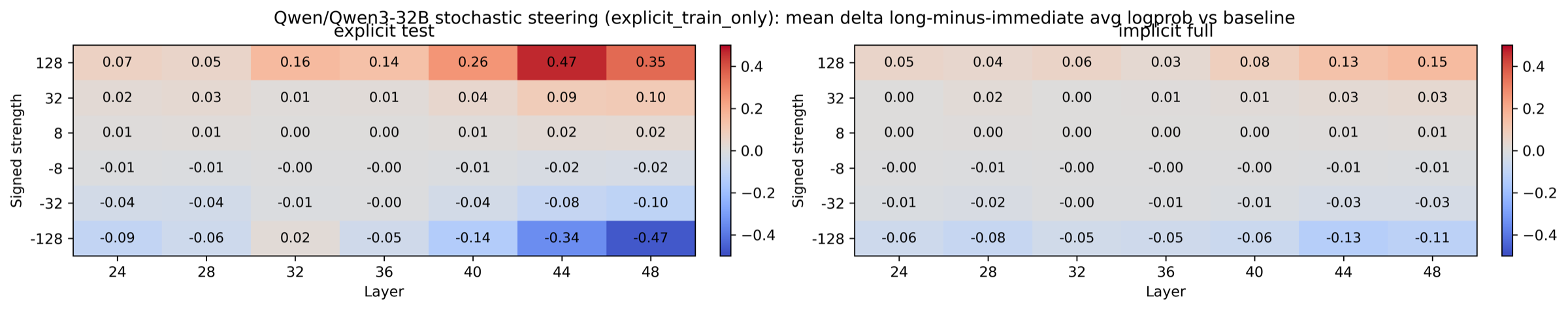}
\includegraphics[width=0.92\textwidth]{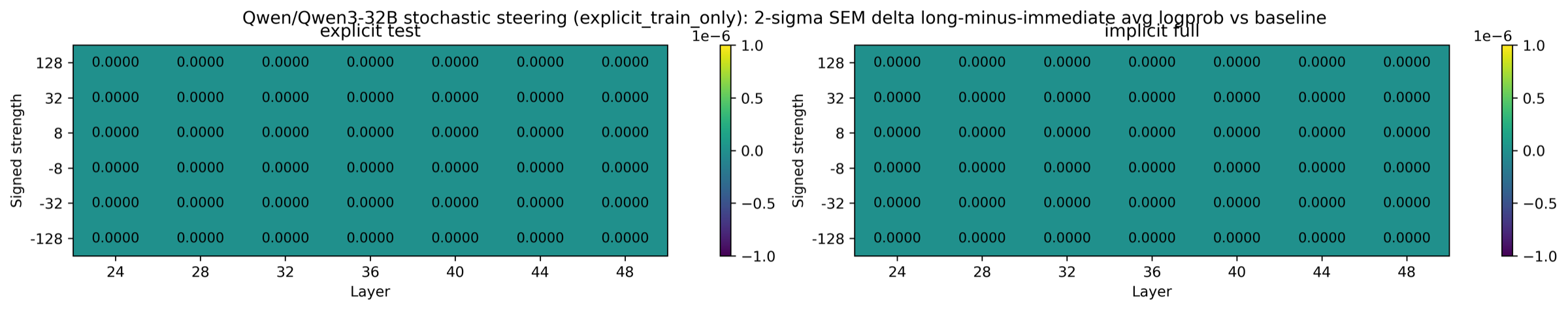}
\caption{Stochastic binary steering forced-choice logprob results. Top: mean
change in long-minus-immediate average logprob margin relative to baseline.
Bottom: 2-sigma standard-error bar across five sampled decodings.}
\label{fig:binary-stochastic-logprob}
\end{figure*}

\begin{figure*}[t]
\centering
\includegraphics[width=0.92\textwidth]{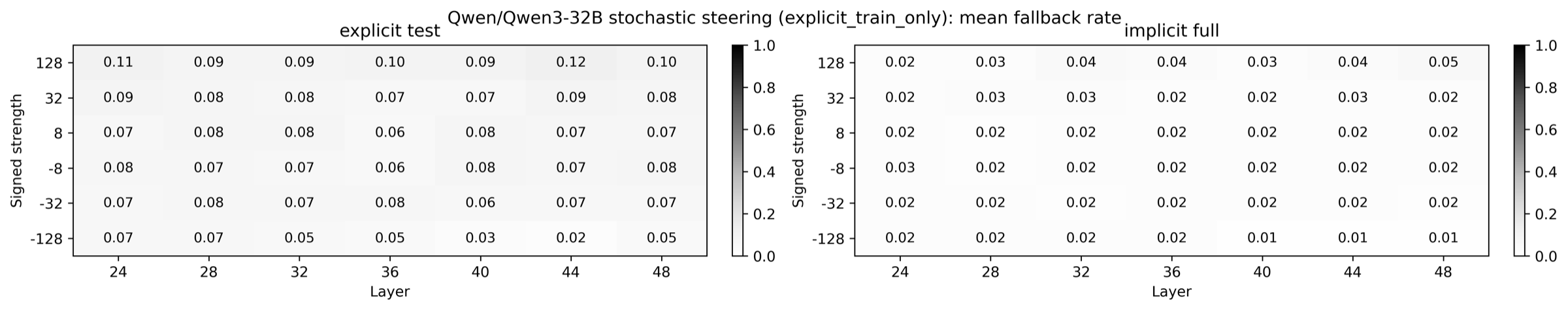}
\includegraphics[width=0.92\textwidth]{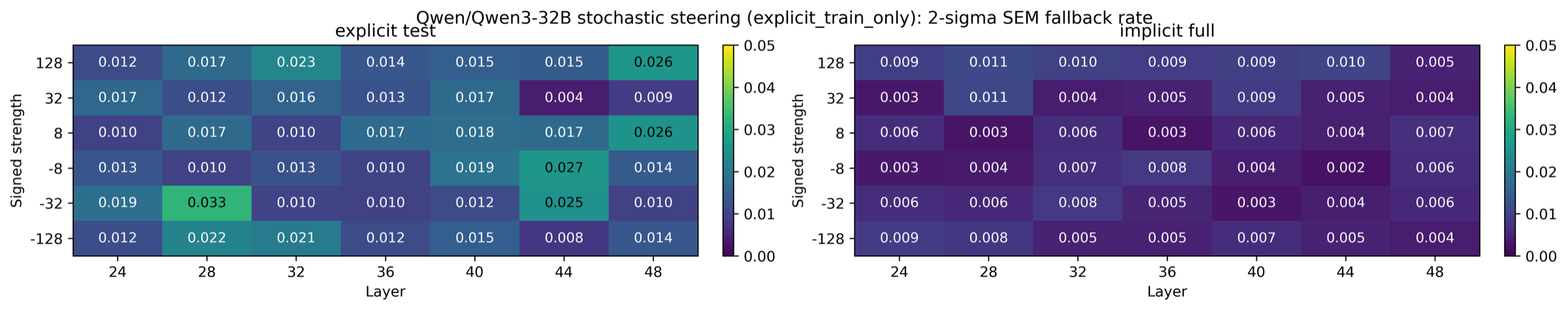}
\caption{Stochastic binary steering fallback diagnostics. Top: mean fallback
rate. Bottom: 2-sigma standard-error bar across five sampled decodings.}
\label{fig:binary-stochastic-fallback}
\end{figure*}

The main binary steering sweep is decoded greedily. To estimate sensitivity to
sampling randomness, Figures~\ref{fig:binary-stochastic-choice}--\ref{fig:binary-stochastic-fallback}
show the stochastic layer sweep with temperature \(0.8\), top-\(p=1\), and five
sampled generations per prompt. In Figures~\ref{fig:binary-stochastic-choice}--\ref{fig:binary-stochastic-fallback}
and Table~\ref{tab:binary-stochastic}, error bars are 2-sigma standard errors,
computed as \(2\sigma/\sqrt{5}\) across the five repeat-level summaries for
each cell. The sweep covers all seven probe layers but only the signed
strengths \(0,\pm8,\pm32,\pm128\) due to compute constraints. The sampled
subset preserves the qualitative steering pattern, suggesting the greedy result
is not an artifact of deterministic decoding.
Table~\ref{tab:binary-stochastic} reports the layer-44 explicit-test slice
corresponding to the main binary steering contrast.

\begin{table*}[t]
\centering
\small
\setlength{\tabcolsep}{5pt}
\begin{tabular}{lccc}
\toprule
Residual addition & Greedy no-fallback long & Sampled no-fallback long & Sampled fallback \\
\midrule
\(-128\hat d_{44}\) & 33.7\% & \(40.6\pm1.9\%\) & \(2.4\pm0.8\%\) \\
0 & 76.8\% & \(74.3\pm1.8\%\) & \(6.6\pm2.2\%\) \\
\(+128\hat d_{44}\) & 93.1\% & \(91.4\pm1.8\%\) & \(12.2\pm1.5\%\) \\
\bottomrule
\end{tabular}
\caption{Stochastic binary steering repeat check on the explicit test set. The
sampled columns report mean \(\pm\) 2-sigma standard-error bars across five
repeat-level sampled decodings.}
\label{tab:binary-stochastic}
\end{table*}

\section{Residual-Norm Diagnostics}
\label{app:residual-norms}

\begin{figure*}[t]
\centering
\includegraphics[width=0.74\textwidth]{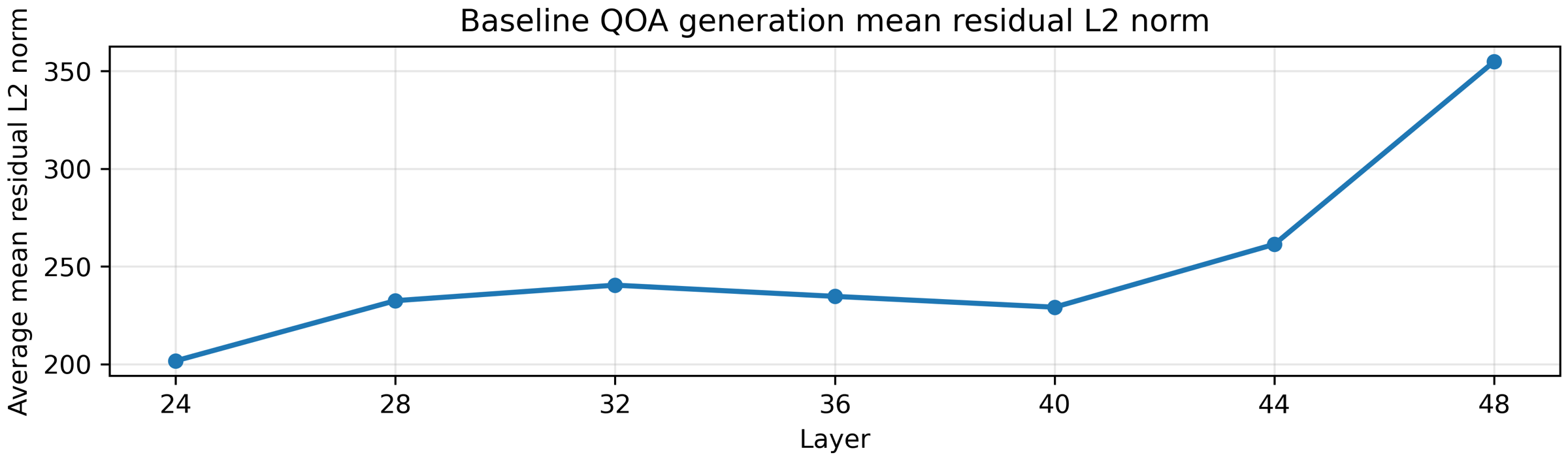}
\caption{Section~\ref{sec:binary-steering} steering experiment hook-site
residual-norm diagnostics from unsteered baseline generations. The plot reports
mean residual-stream L2 norm at the prompt/decode positions where steering is
applied.}
\label{fig:binary-hook-residual-norms}
\end{figure*}

\begin{figure*}[t]
\centering
\includegraphics[width=0.74\textwidth]{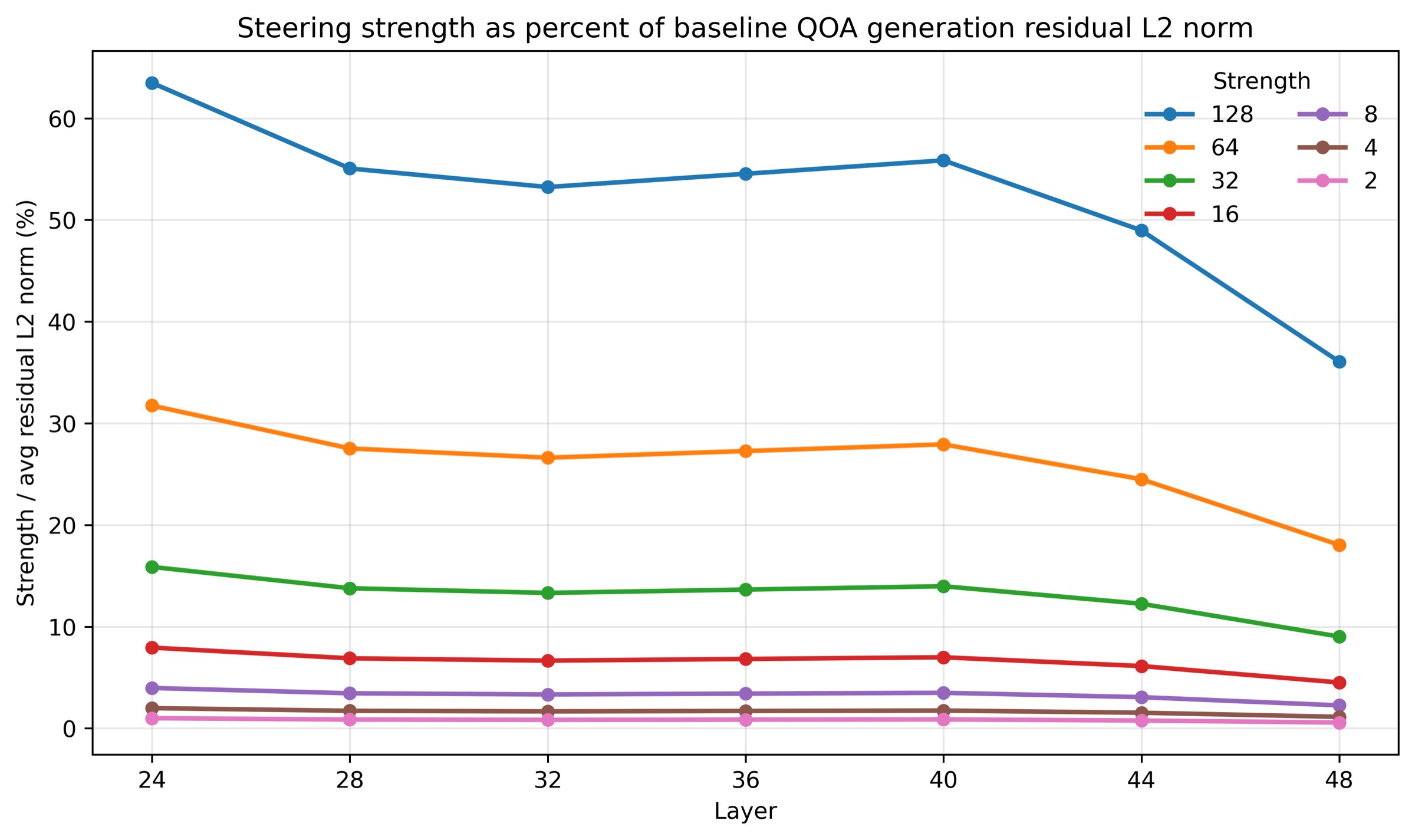}
\caption{Section~\ref{sec:binary-steering} steering experiment intervention
scale. The plot reports steering strength as a percentage of the mean
hook-site residual norm from unsteered baseline generations.}
\label{fig:binary-hook-steering-scale}
\end{figure*}

\begin{figure*}[t]
\centering
\includegraphics[width=0.74\textwidth]{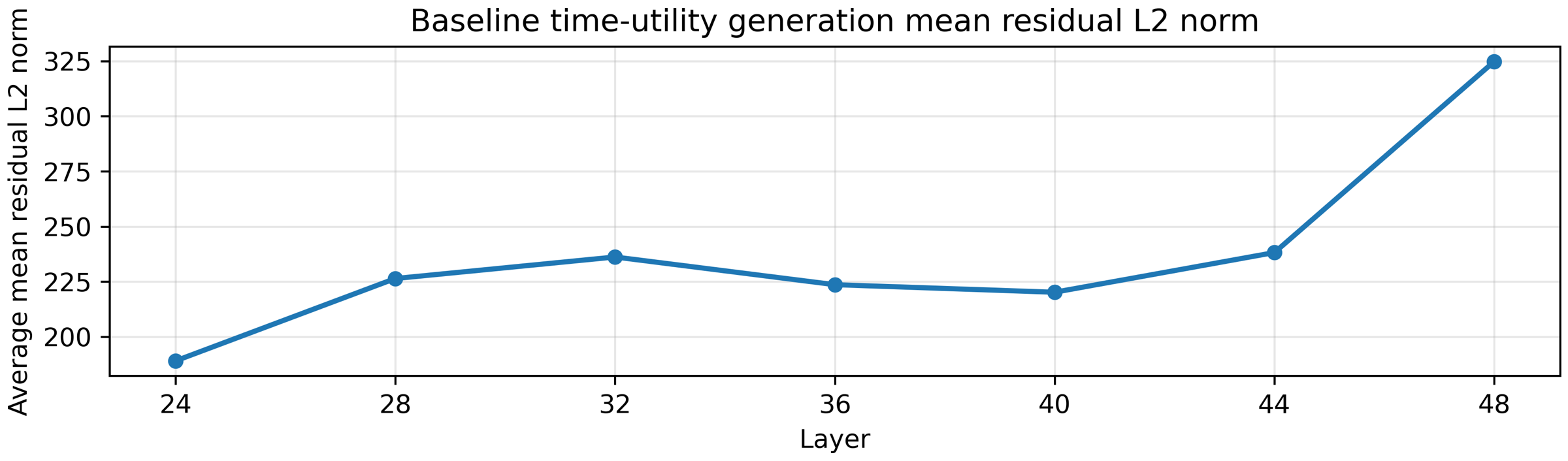}
\caption{Section~\ref{sec:time-utility} steering experiment hook-site
residual-norm diagnostics from unsteered baseline generations. The plot reports
mean residual-stream L2 norm at the prompt/decode positions where steering is
applied.}
\label{fig:time-hook-residual-norms}
\end{figure*}

\begin{figure*}[t]
\centering
\includegraphics[width=0.74\textwidth]{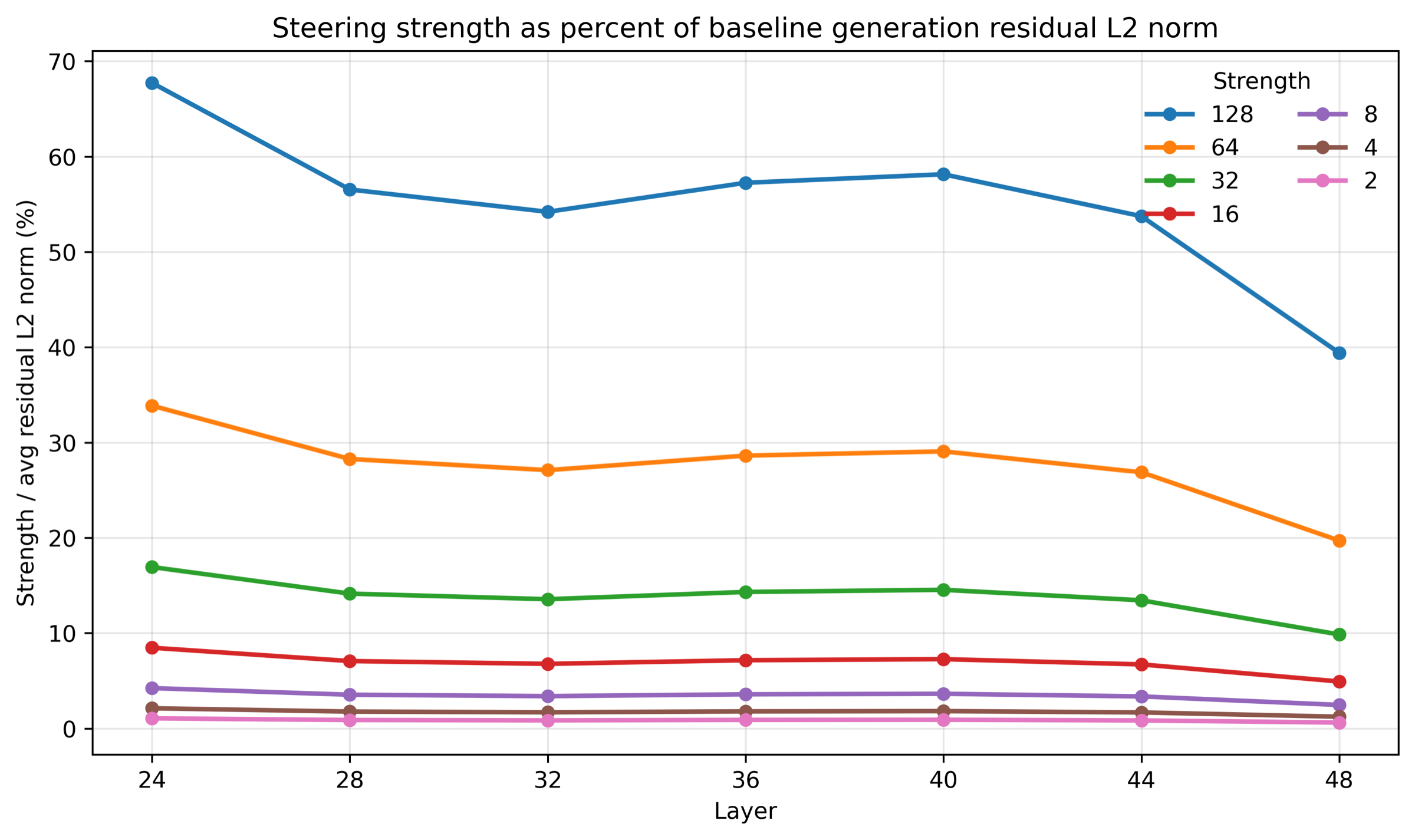}
\caption{Section~\ref{sec:time-utility} steering experiment intervention
scale. The plot reports steering strength as a percentage of the mean hook-site
residual norm from unsteered baseline generations.}
\label{fig:time-hook-steering-scale}
\end{figure*}

\section{Time-Utility Diagnostics}
\label{app:time-utility-details}

\begin{figure*}[t]
\centering
\includegraphics[width=0.98\textwidth,height=0.88\textheight,keepaspectratio]{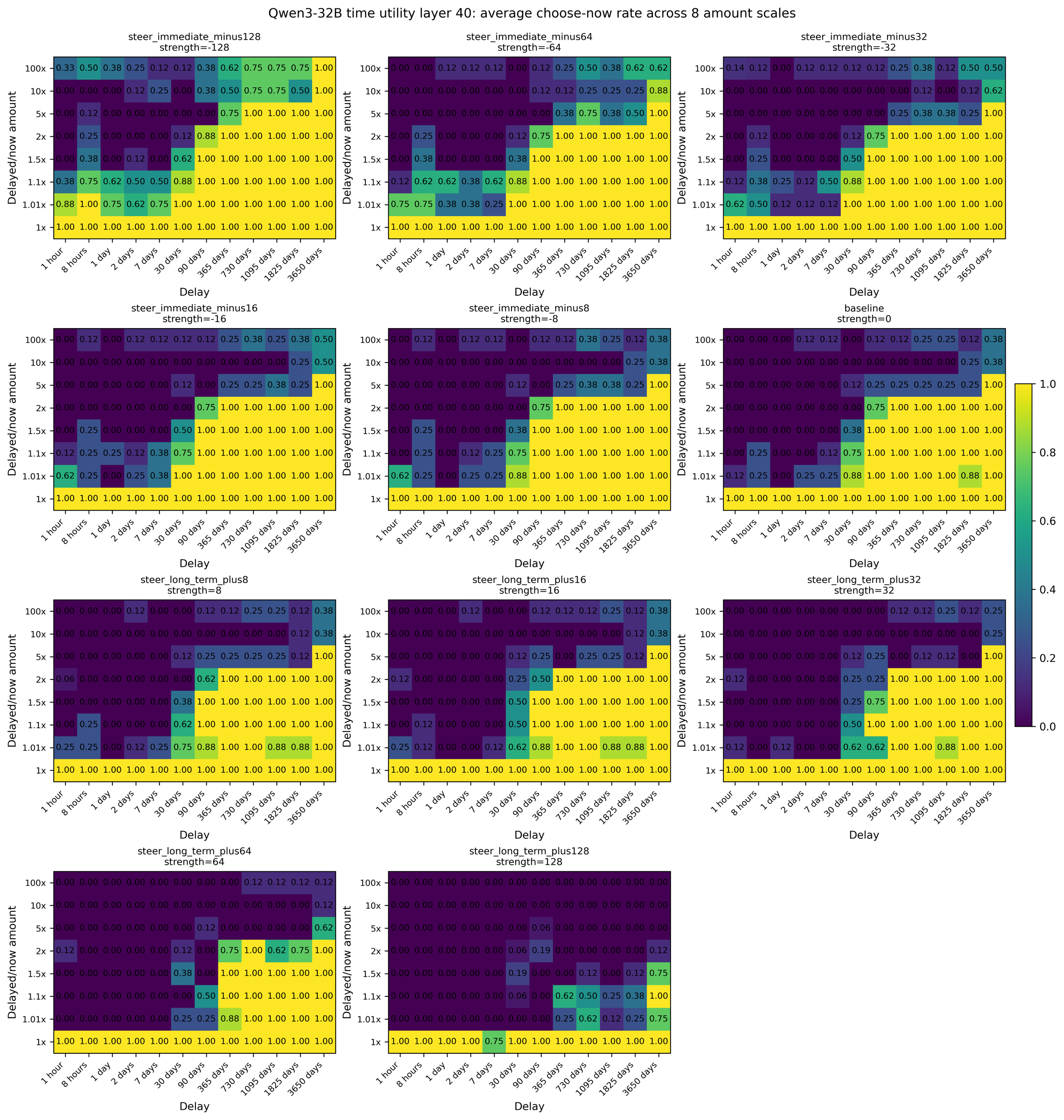}
\caption{Average present-choice heatmaps for the layer-40 time-utility
experiment. Each panel averages over amount scale factors. Columns are delays,
rows are delayed/present amount multipliers, and color is the parsed fraction
choosing the present amount.}
\label{fig:time-utility-present-heatmaps}
\end{figure*}

Figure~\ref{fig:time-utility-present-heatmaps} shows the full grid underlying
the AEIM summaries. Yellow cells are high present-choice proportions, and dark
cells are low present-choice proportions. Because delay increases from left to right and delayed/present
amount increases from bottom to top, a rationally expected pattern has a yellow
cluster in the bottom-right region: when the delayed amount is not much larger than
the present amount and the wait is long, the present option should usually be
preferred. Conversely, the top-left region should be dark, because a much
larger delayed reward after a short wait should usually be preferred. The
steering pattern follows the intended direction: negative strengths expand the
yellow present-choice region, while positive strengths contract it; the
expansion and contraction become larger as steering strength increases.

\begin{figure*}[t]
\centering
\includegraphics[width=0.98\textwidth]{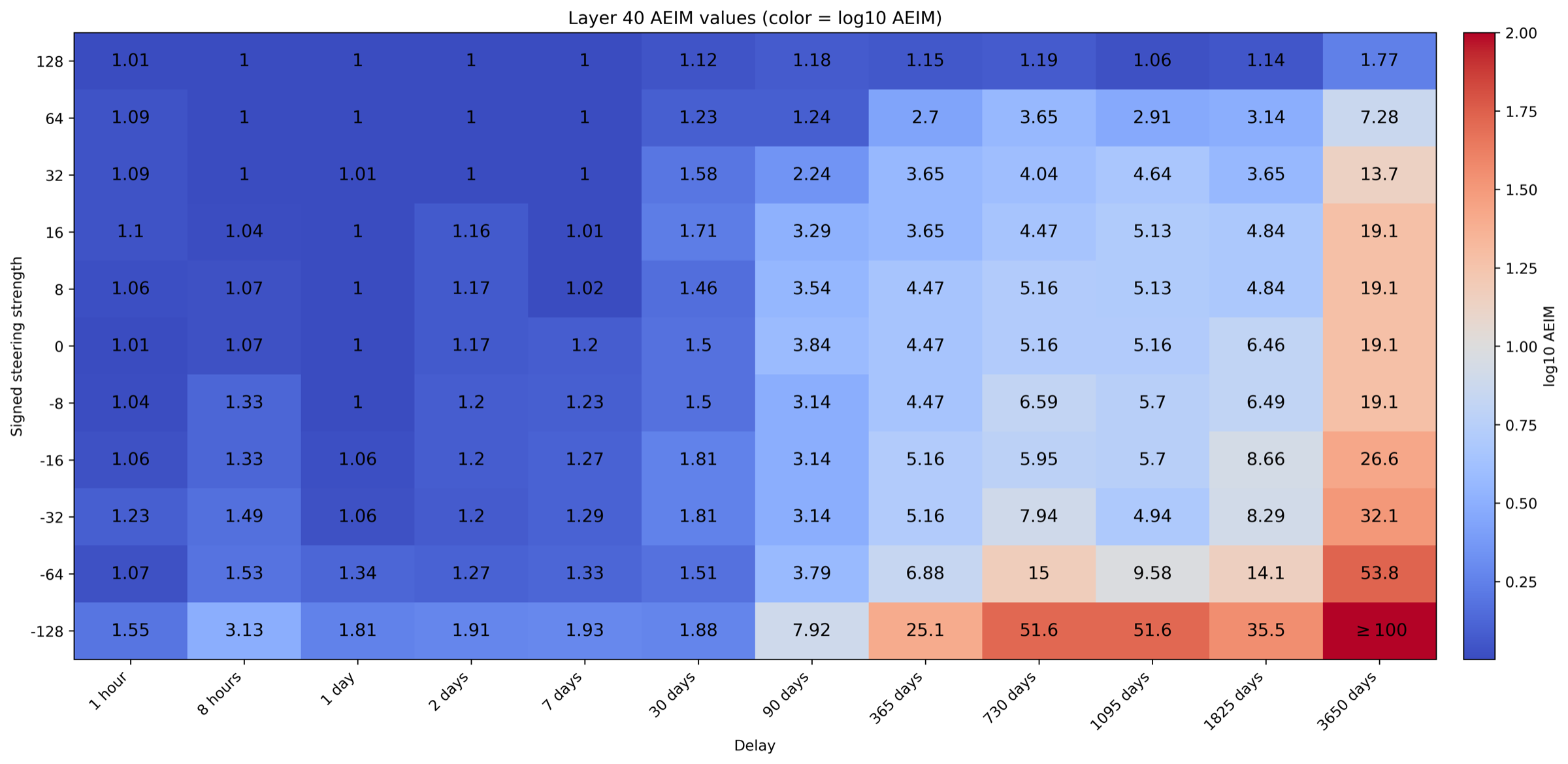}
\caption{Layer-40 area-equivalent indifference multipliers for all delays and
steering strengths. Higher values mean a larger delayed amount is needed to
offset the delay. The \(-128\), 3650-day point is right-censored at the largest
tested multiplier because all valid parsed rows choose the immediate option; the
true value is therefore \(\ge100\).}
\label{fig:time-utility-aeim-values}
\end{figure*}

\begin{figure*}[t]
\centering
\includegraphics[width=0.48\textwidth]{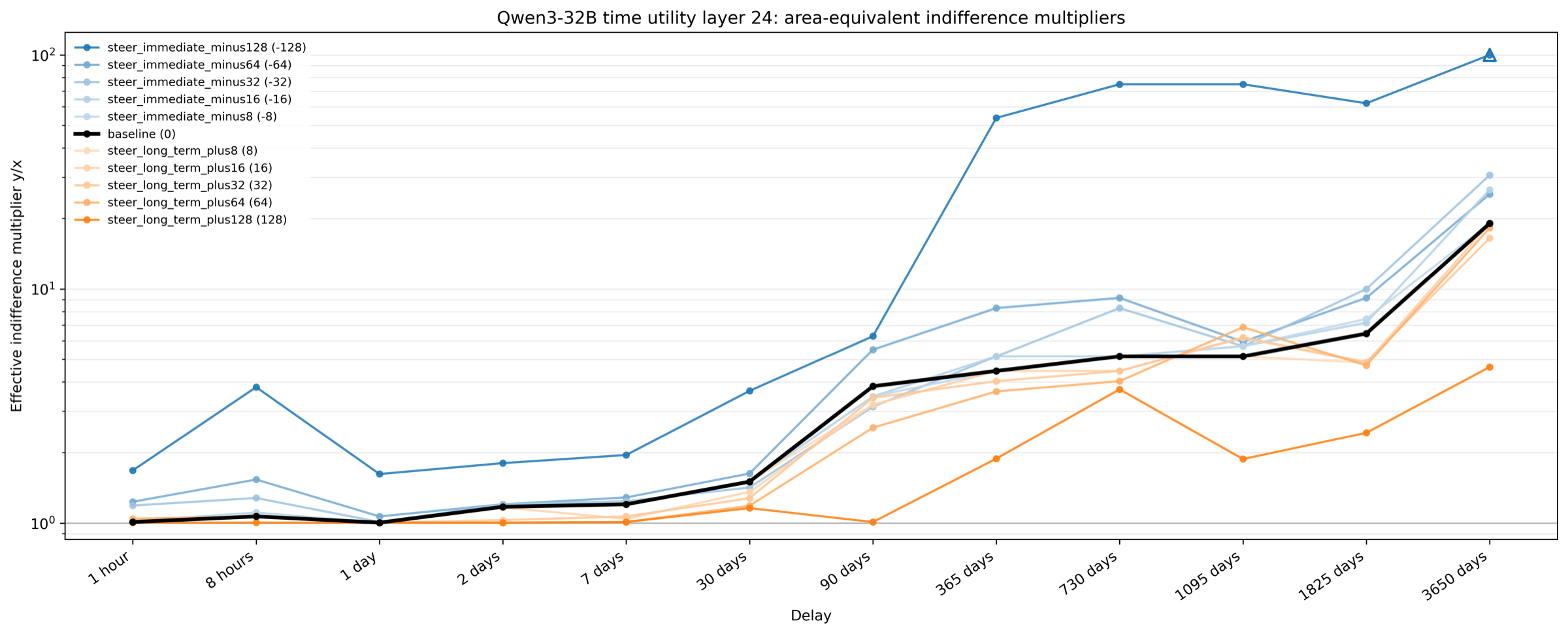}
\includegraphics[width=0.48\textwidth]{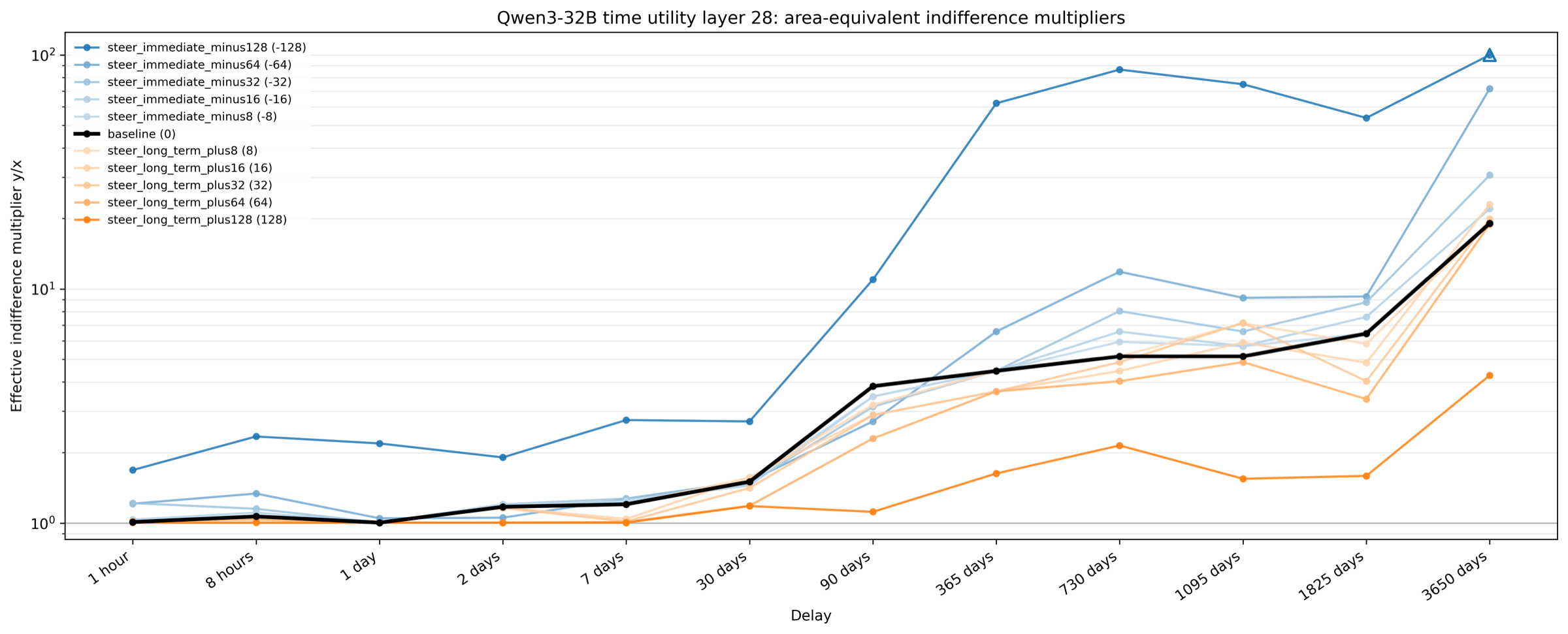}
\includegraphics[width=0.48\textwidth]{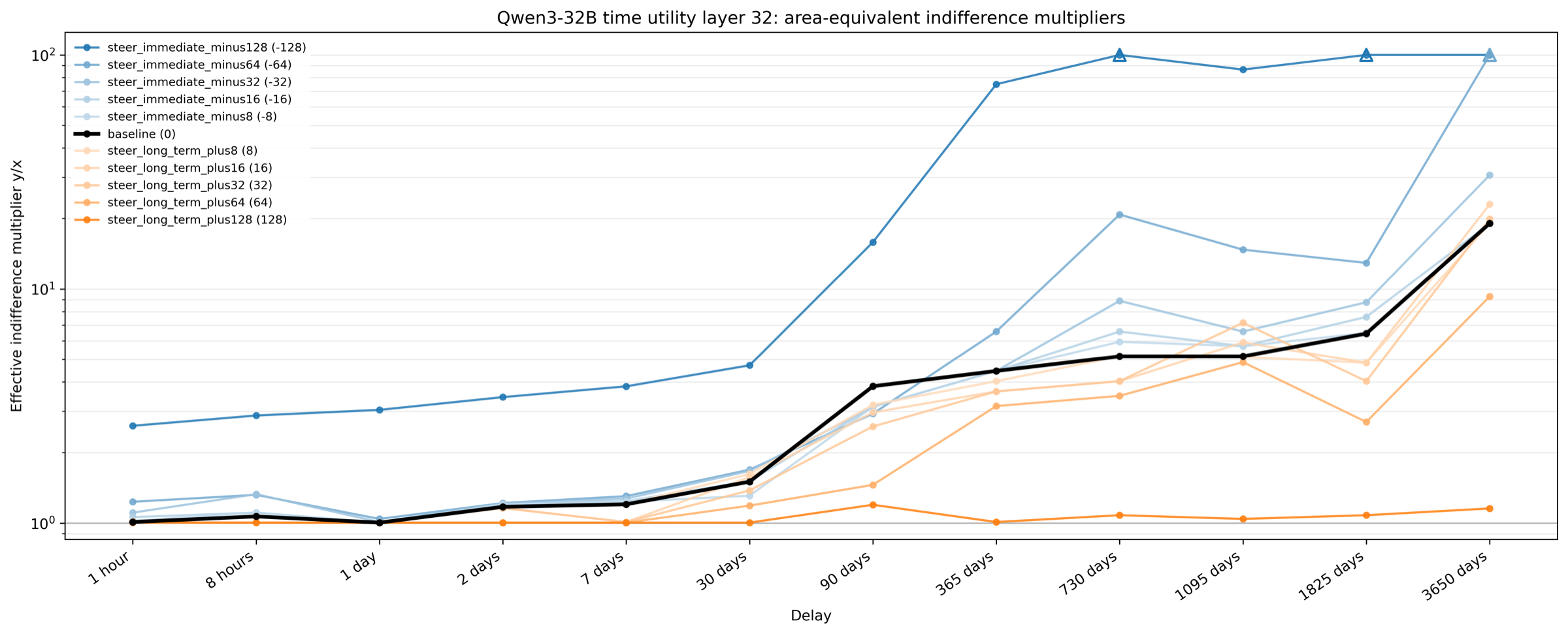}
\includegraphics[width=0.48\textwidth]{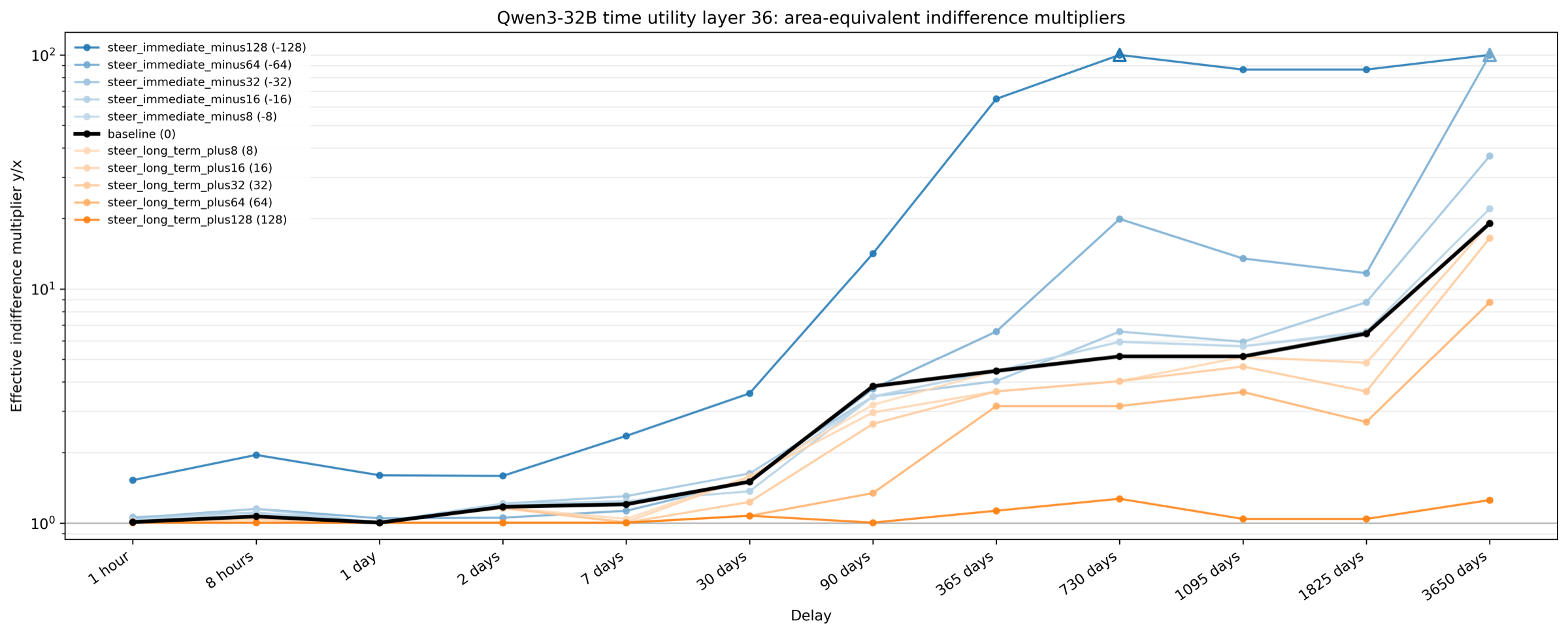}
\includegraphics[width=0.48\textwidth]{qwen3_32b_time_utility_layer_sweep_layer40_aeim_curves.png}
\includegraphics[width=0.48\textwidth]{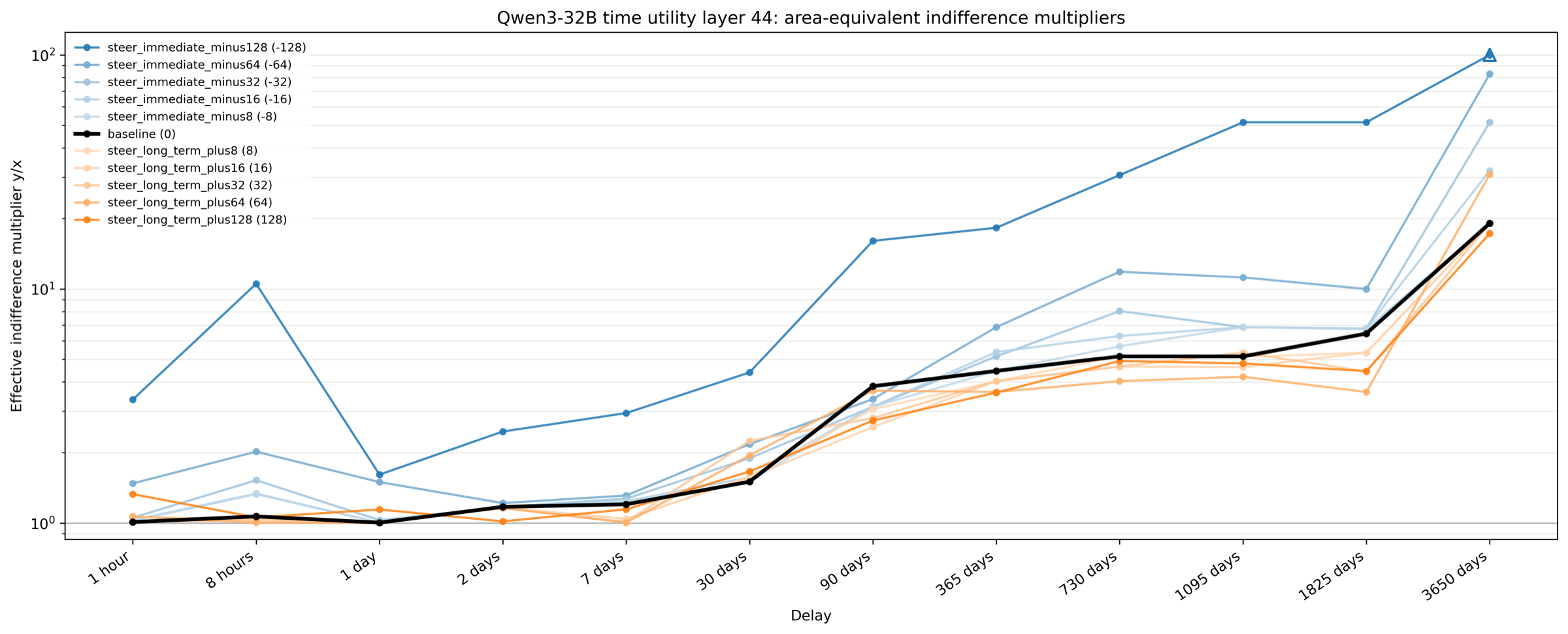}
\includegraphics[width=0.48\textwidth]{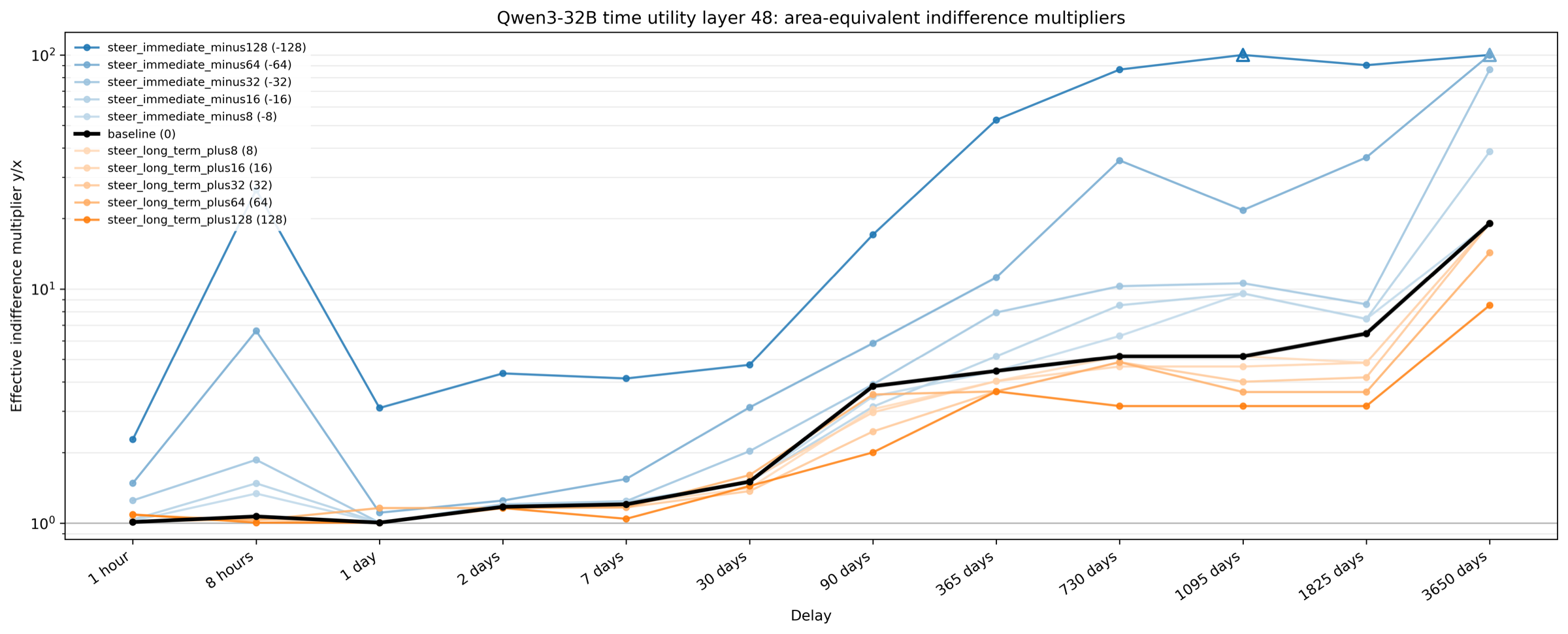}
\caption{Layerwise time-utility AEIM curves under CAA steering with the MM
difference direction. These
are the per-layer curves summarized by Figure~\ref{fig:time-utility-layer-sweep}.}
\label{fig:time-utility-all-layer-aeim}
\end{figure*}

\begin{figure*}[t]
\centering
\includegraphics[width=0.88\textwidth]{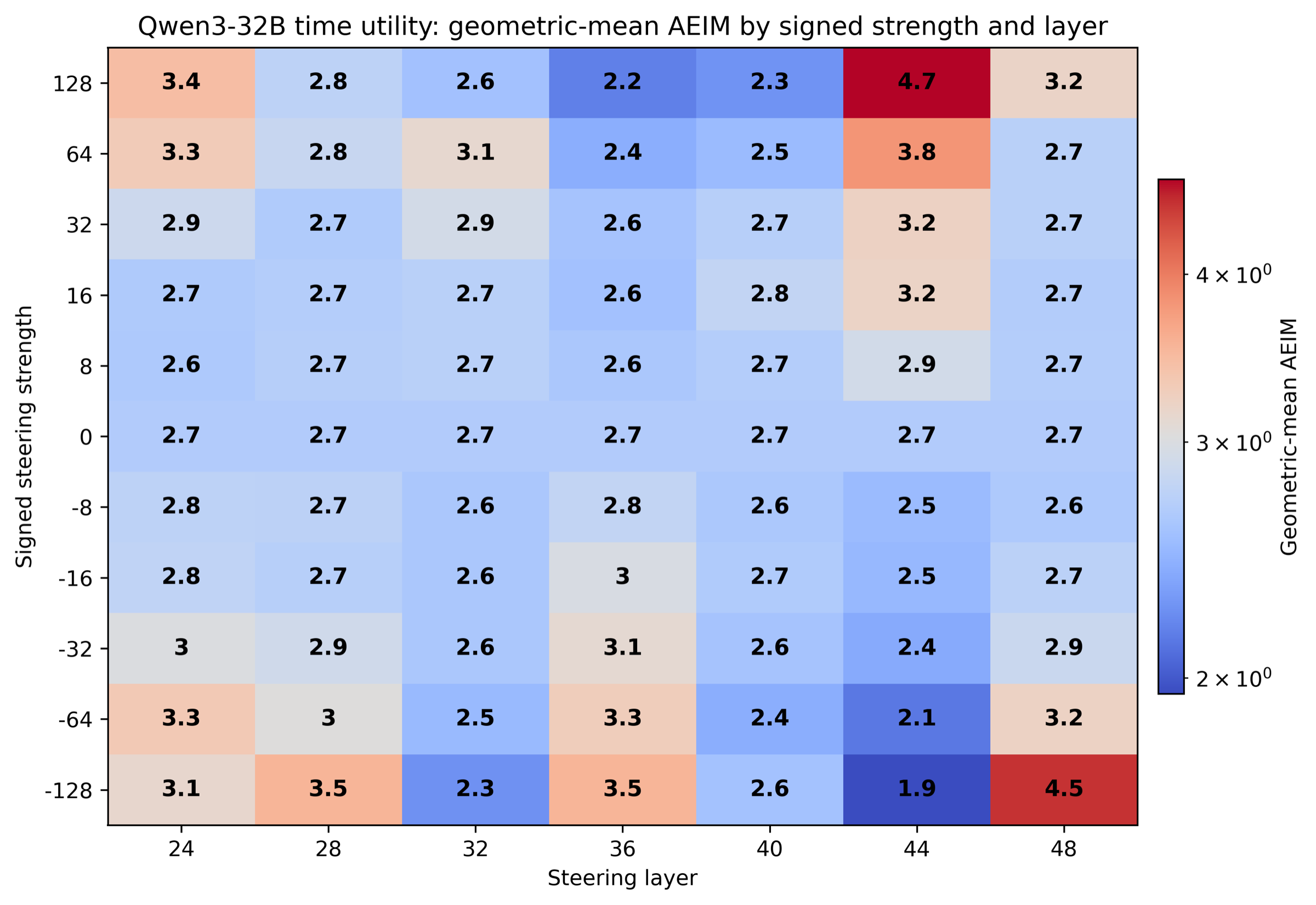}
\caption{Time-utility random-orthogonal control. The control uses matched-norm
unit vectors orthogonal to the MM difference vector at each layer. It produces some AEIM
movement, but the direction is not consistently aligned with steering sign:
at \(\pm128\), the negative-over-positive geometric-mean AEIM ratio ranges from
0.39 to 1.69 across layers, compared with 4.55 to 14.46 for CAA steering with
the MM difference direction.}
\label{fig:time-utility-random-orthogonal-control}
\end{figure*}

\section{Stochastic Time-Utility Robustness}
\label{app:time-utility-stochastic}

\begin{figure*}[t]
\centering
\includegraphics[width=0.68\textwidth]{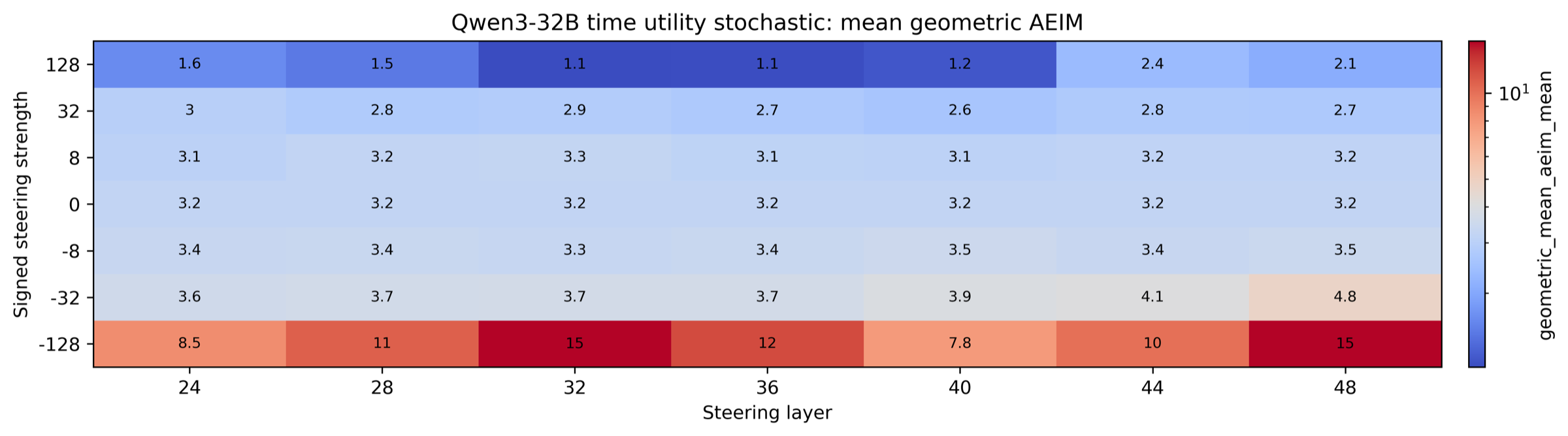}
\includegraphics[width=0.68\textwidth]{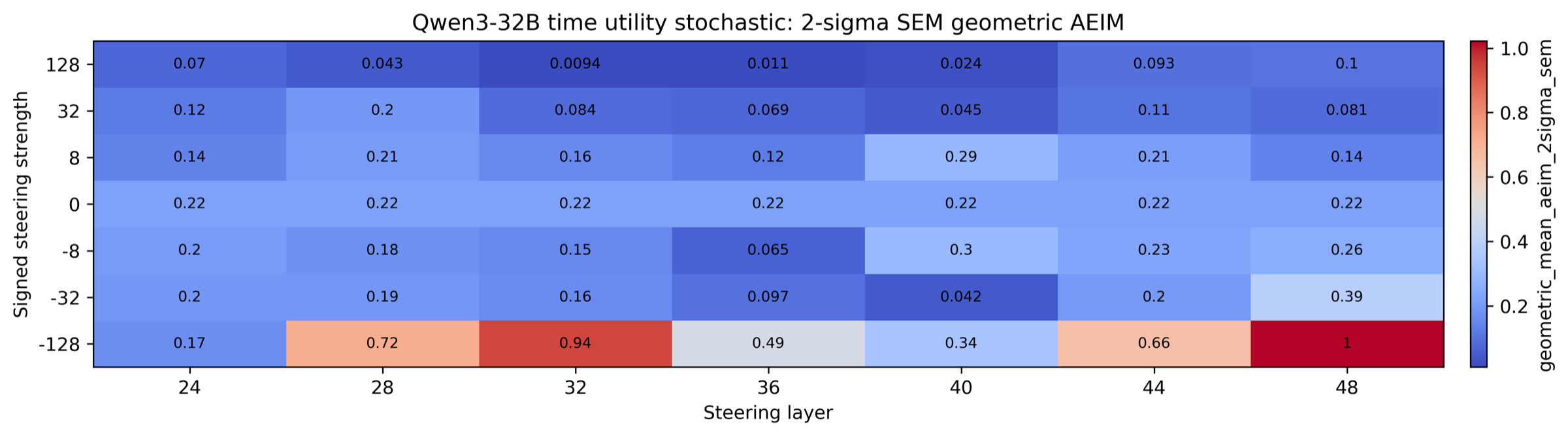}
\caption{Stochastic time-utility geometric-mean AEIM layer sweep. Top: mean
geometric-mean AEIM across delays. Bottom: 2-sigma standard-error bar across
five sampled decodings.}
\label{fig:time-utility-stochastic-geomean}
\end{figure*}

\begin{figure*}[t]
\centering
\includegraphics[width=0.48\textwidth]{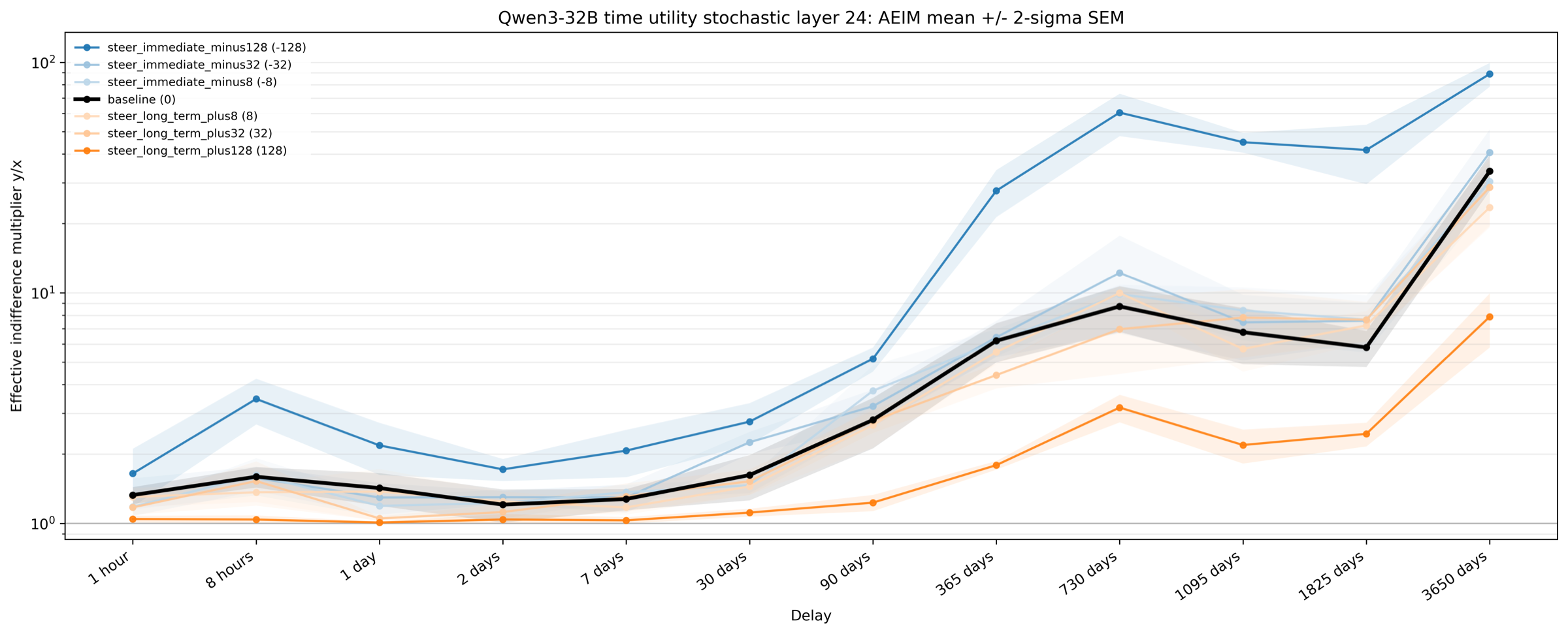}
\includegraphics[width=0.48\textwidth]{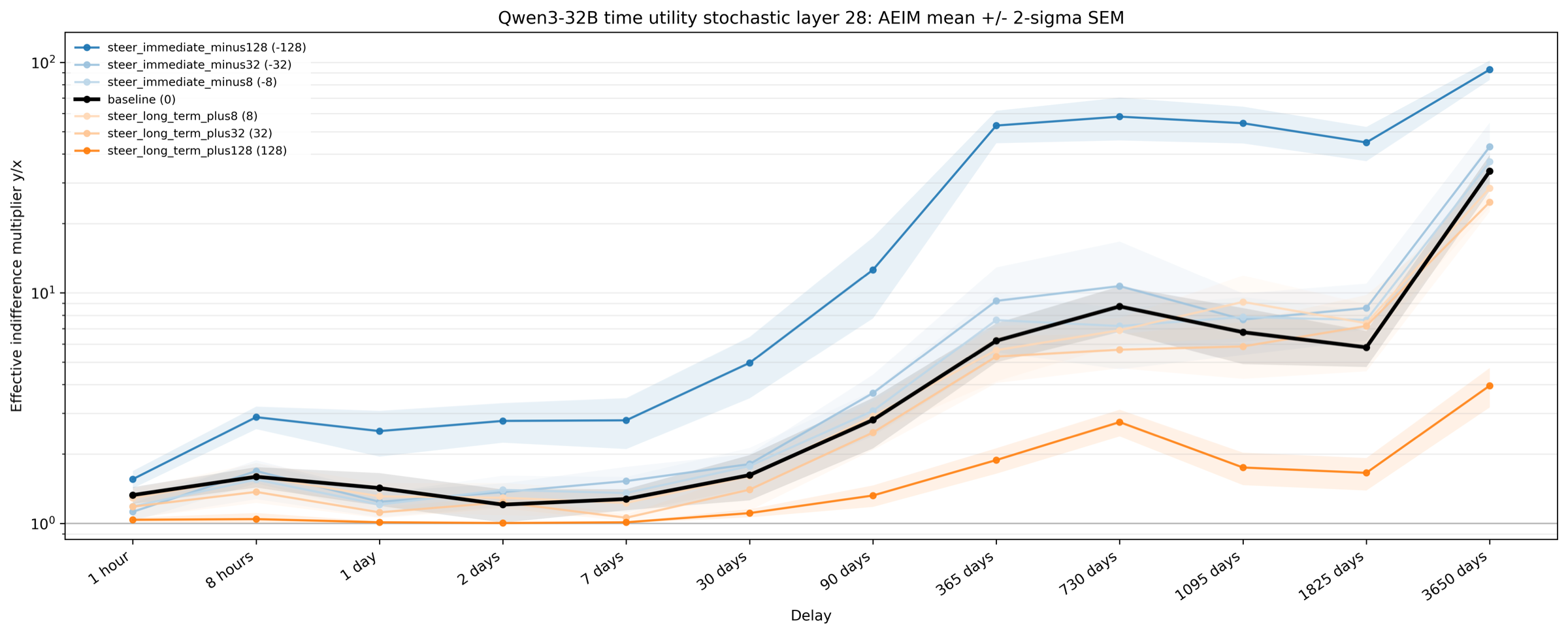}
\includegraphics[width=0.48\textwidth]{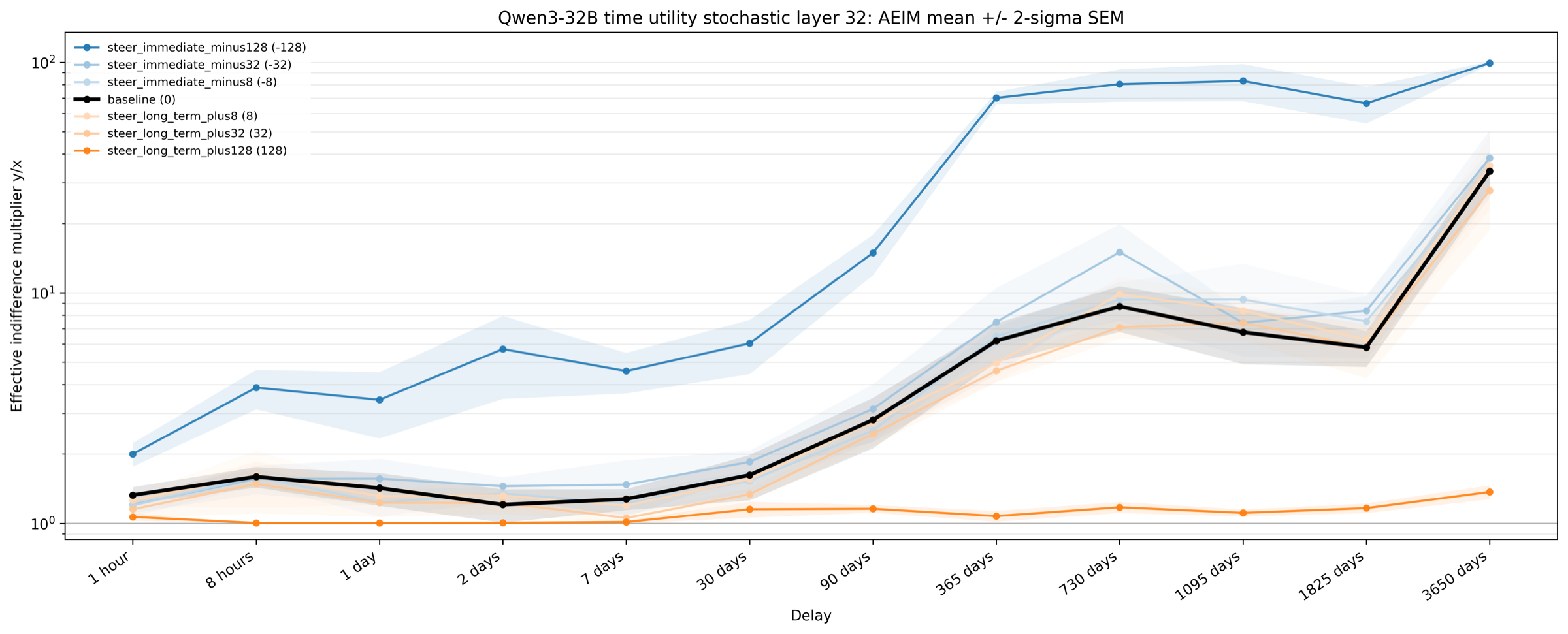}
\includegraphics[width=0.48\textwidth]{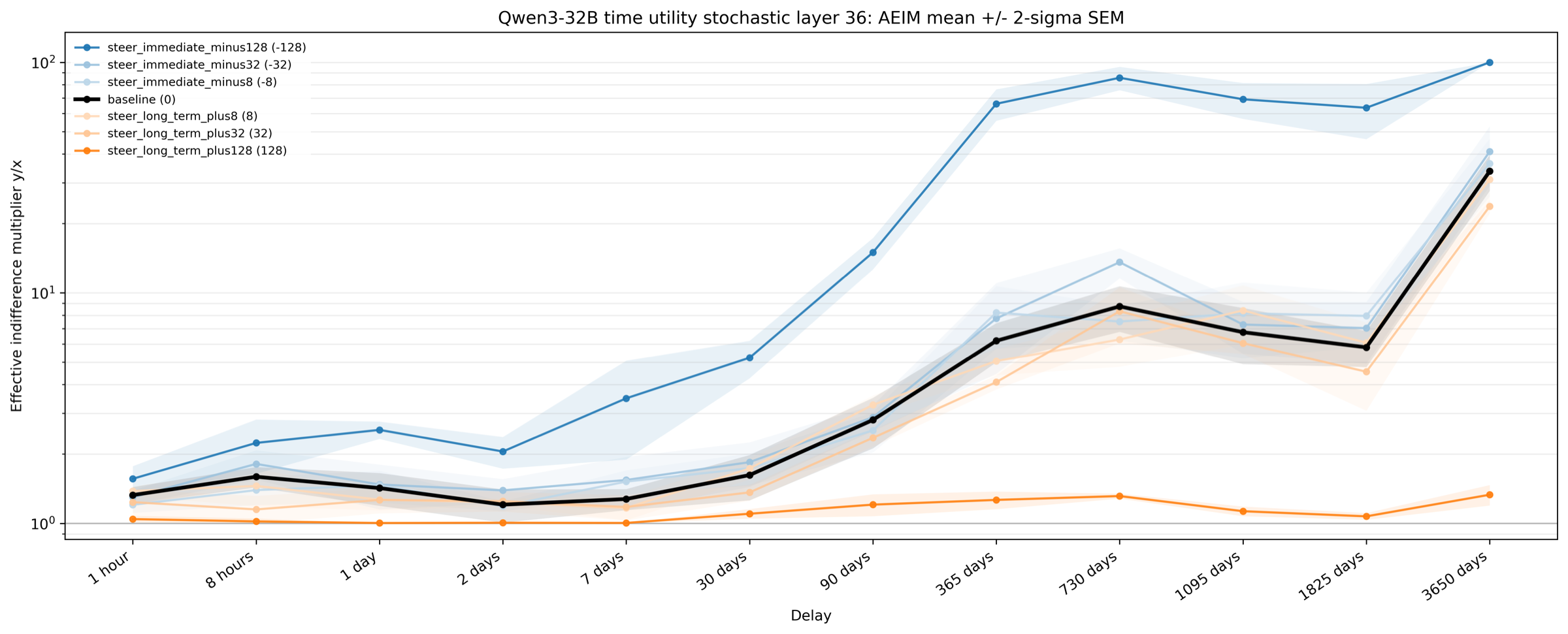}
\includegraphics[width=0.48\textwidth]{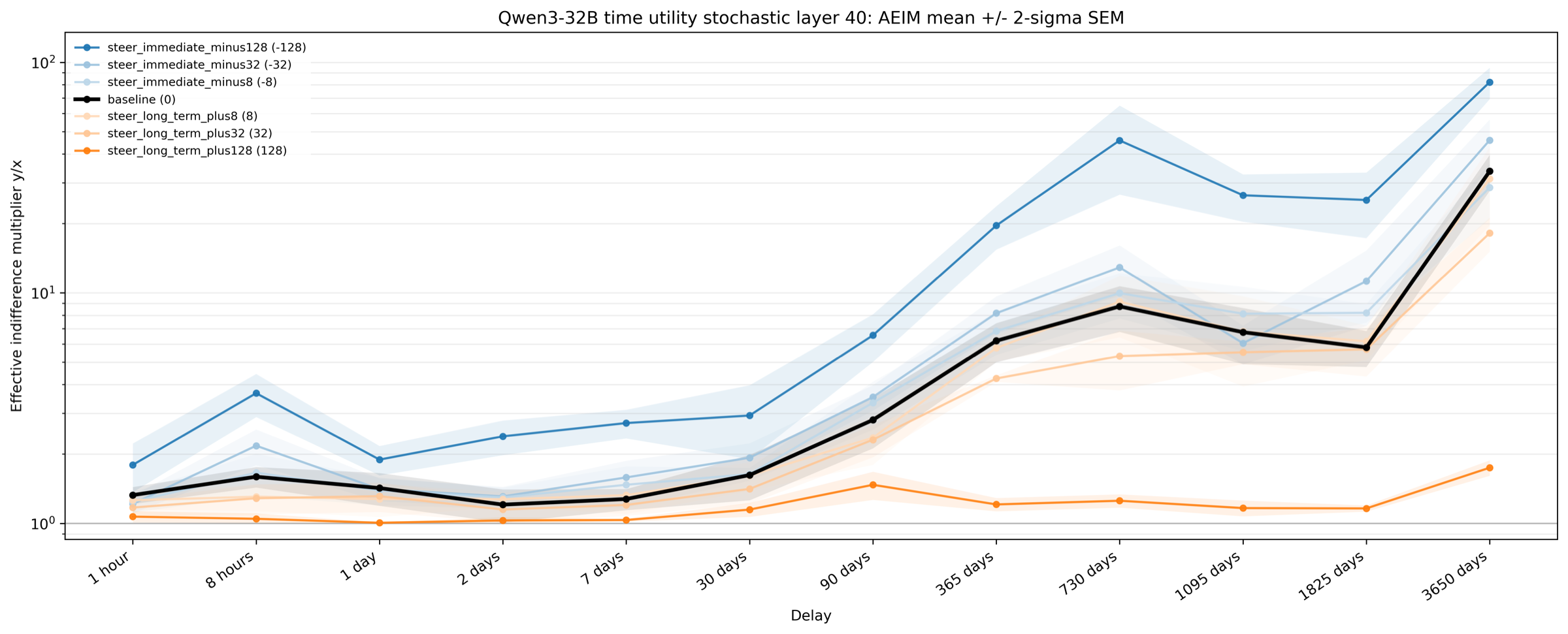}
\includegraphics[width=0.48\textwidth]{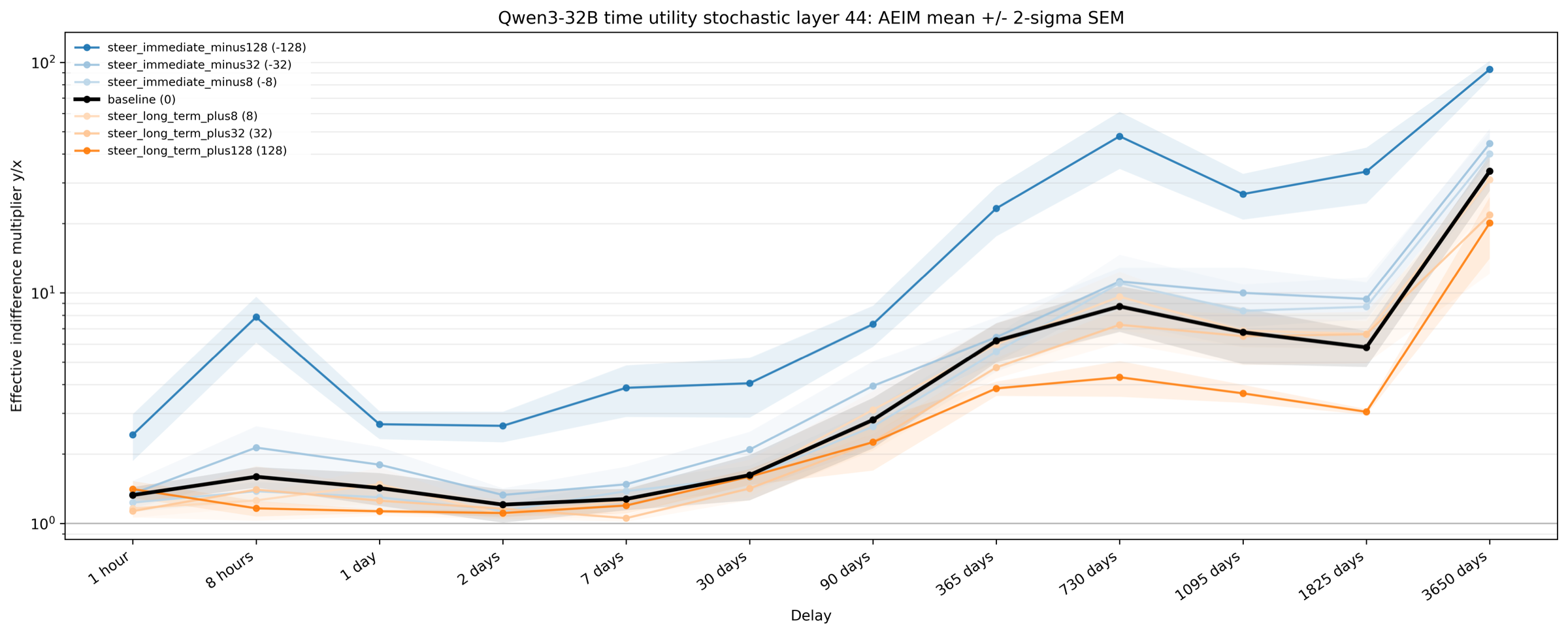}
\includegraphics[width=0.48\textwidth]{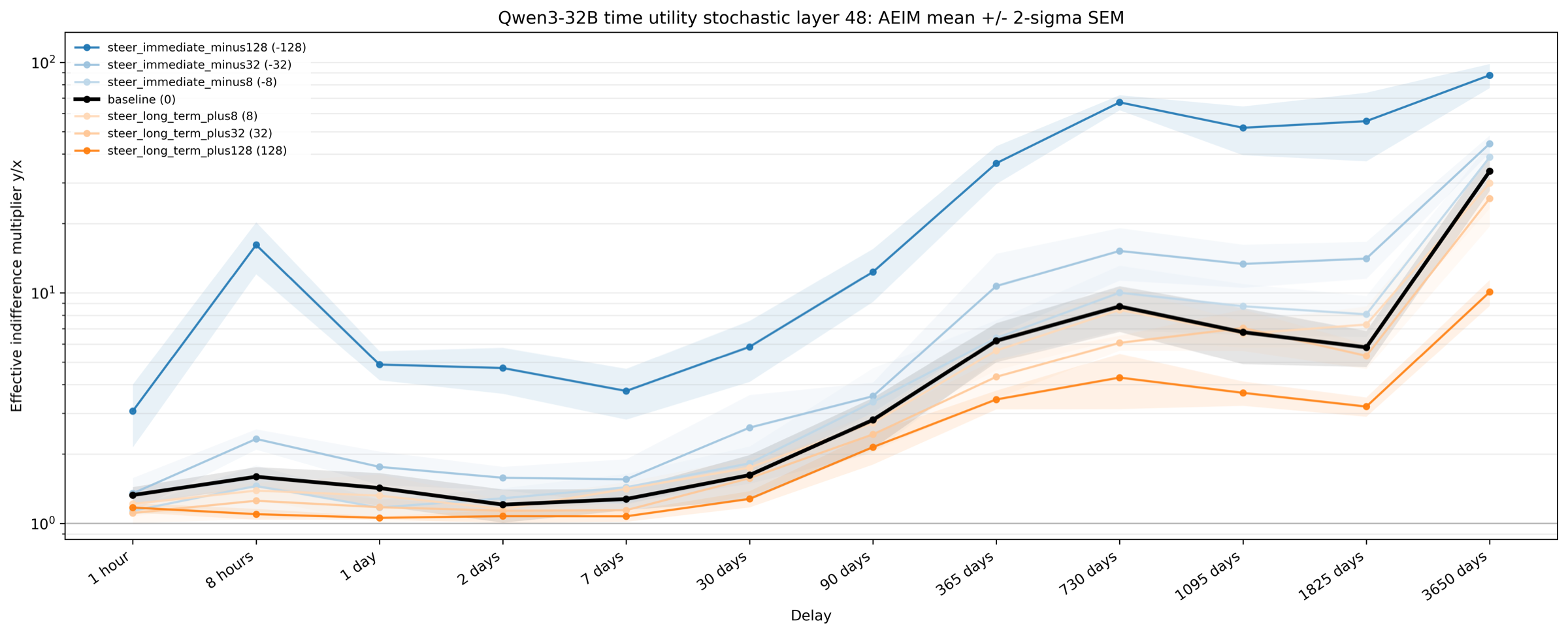}
\caption{Stochastic time-utility AEIM curves for all seven probe layers. Bands show
2-sigma standard-error bars across five sampled decodings per prompt at
temperature \(0.8\) and top-\(p=1\).}
\label{fig:time-utility-stochastic-aeim}
\end{figure*}

The main time-utility sweep is decoded greedily. To estimate sensitivity to
sampling randomness, Figures~\ref{fig:time-utility-stochastic-geomean} and
\ref{fig:time-utility-stochastic-aeim} repeat a stochastic layer sweep with
temperature \(0.8\), top-\(p=1\), and five sampled generations per prompt. The
sweep covers all seven probe layers at signed strengths \(0,\pm8,\pm32,\pm128\),
a strength subset of the full greedy grid in the main text due to compute constraints. This stochastic
sweep preserves the qualitative steering pattern, suggesting the greedy result
is not an artifact of deterministic decoding. In
Figures~\ref{fig:time-utility-stochastic-geomean}--\ref{fig:time-utility-stochastic-aeim}
and Table~\ref{tab:time-utility-stochastic}, error bars are 2-sigma standard
errors, computed as \(2\sigma/\sqrt{5}\) across the five repeat-level summaries
for each cell.

\begin{table*}[t]
\centering
\small
\setlength{\tabcolsep}{5pt}
\begin{tabular}{lccc}
\toprule
Layer & \(-128\hat d_l\) & 0 & \(+128\hat d_l\) \\
\midrule
24 & \(8.48\pm0.17\) & \(3.22\pm0.22\) & \(1.63\pm0.07\) \\
28 & \(10.95\pm0.72\) & \(3.22\pm0.22\) & \(1.46\pm0.04\) \\
32 & \(15.20\pm0.94\) & \(3.22\pm0.22\) & \(1.10\pm0.01\) \\
36 & \(12.08\pm0.49\) & \(3.22\pm0.22\) & \(1.12\pm0.01\) \\
40 & \(7.84\pm0.34\) & \(3.22\pm0.22\) & \(1.18\pm0.02\) \\
44 & \(10.08\pm0.66\) & \(3.22\pm0.22\) & \(2.36\pm0.09\) \\
48 & \(15.11\pm1.02\) & \(3.22\pm0.22\) & \(2.09\pm0.10\) \\
\bottomrule
\end{tabular}
\caption{Stochastic time-utility geometric-mean AEIM by layer. Entries report
mean \(\pm\) 2-sigma standard-error bars across five repeat-level sampled
decodings.}
\label{tab:time-utility-stochastic}
\end{table*}

\FloatBarrier

\end{document}